\documentclass[11pt,a4paper]{article}
\usepackage[hyperref]{acl2021}
\usepackage{times}
\usepackage{latexsym}

\usepackage{amsmath}
\usepackage{graphicx}
\usepackage{float}
\usepackage{subcaption}
\usepackage{multicol}
\usepackage{booktabs}
\usepackage{multirow}
\usepackage[bottom]{footmisc}
\usepackage{cancel}
\usepackage{caption}
\usepackage{subcaption}
\usepackage{pifont}
\newcommand{\cmark}{\ding{51}}%
\newcommand{\xmark}{\ding{55}}%
\DeclareUnicodeCharacter{0307}{} 

\usepackage{microtype}

\aclfinalcopy 
\def\aclpaperid{487} 

\title{Long-Span Summarization via Local Attention and Content Selection}

\author{Potsawee Manakul \textnormal{ and } Mark J. F. Gales \\
  Department of Engineering, University of Cambridge \\
  \texttt{pm574@cam.ac.uk, mjfg@eng.cam.ac.uk}}

\date{}

\begin{document}
\maketitle
\begin{abstract}

Transformer-based models have achieved state-of-the-art results in a wide range of natural language processing (NLP) tasks including document summarization. Typically these systems are trained by fine-tuning a large pre-trained model to the target task. One issue with these transformer-based models is that they do not scale well in terms of memory and compute requirements as the input length grows. Thus, for long document summarization, it can be challenging to train or fine-tune these models. In this work, we exploit large pre-trained transformer-based models and address long-span dependencies in abstractive summarization using two methods: local self-attention; and explicit content selection. These approaches are compared on a range of network configurations. Experiments are carried out on standard long-span summarization tasks, including Spotify Podcast, arXiv, and PubMed datasets. We demonstrate that by combining these methods, we can achieve state-of-the-art results on all three tasks in the ROUGE scores. Moreover, without a large-scale GPU card, our approach can achieve comparable or better results than existing approaches.\footnote{Our code is available at \url{https://github.com/potsawee/longsum0}.}

\end{abstract}

\section{Introduction}
Transformer-based models \cite{vaswani2017attention} are ubiquitously state-of-art across many natural language processing (NLP) tasks, including summarization. To achieve the best results, the community has trained ever larger transformer models on larger amount of data, and/or more task-specific optimization objectives \cite{devlin2018bert, raffel2020exploring, lewis-etal-2020-bart, brown2020language}. In long document summarization, the input sequences could be more than an order of magnitude longer than the limits of these transformer models. Although the limits can be extended, training large transformer models on long sequences is expensive and may not be possible on a standard GPU card because of the self-attention mechanism that grows quadratically with sequence length.

To tackle the quadratic characteristic, recent works have modified self-attention mechanism and proposed variants of the transformer such that the quadratic complexity is reduced \cite{tay2020efficient, kitaev2020reformer, child2019generating, beltagy2020longformer,ainslie-etal-2020-etc,zaheer2020big}. However, pre-trained weights of the modified models are not readily available. In contrast, standard models such as BERT \cite{devlin2018bert} or BART \cite{lewis-etal-2020-bart} have been trained on various target tasks, including text summarization \cite{liu-lapata-2019-text}. This allows practitioners to achieve good performance with less training time. Thus, we are interested in exploiting pre-trained models for long-span summarization tasks.

We study a range of design configurations empirically and theoretically in regards to memory and compute requirements as well as their performance. We propose that long-span dependencies can be handled by two complementary methods. Firstly, inspired by modified self-attention transformers, we exploit standard transformer models by constraining attention mechanism to be local, allowing longer input spans during training. Secondly, because abstractive summarization systems perform content selection implicitly \cite{nallapati-etal-2016-abstractive, lebanoff-etal-2020-cascade}, to reduce memory and compute requirements an alternative method is to perform content selection explicitly before the abstractive stage. We study content selection during two phases: training time and test time. At training time, we investigate methods to select data for training fixed-span abstractive models. At test time, we extend existing model-based selection methods, and we propose a multitask content selection method that ranks sentences through extractive labelling based module \cite{cheng-lapata-2016-neural} and attention based module \cite{see-etal-2017-get}. Ultimately, we explore the combined approach, consisting of local self-attention transformer and content selection for long-document summarization.

We conduct our experiments using a number of design configurations on the Spotify open-domain Podcast summarization dataset \cite{clifton-etal-2020-100000}. This dataset is challenging not only because of its long-span nature, but also because transcribed spoken utterances typically have lower information density \cite{li-etal-2019-keep, manakul2020_interspeech}. Furthermore, we carry out experiments on arXiv and PubMed datasets \cite{cohan-etal-2018-discourse} to further demonstrate and verify the effectiveness of our approach as well as making comparisons to existing approaches. We highlight the strengths and weaknesses of our approach in different resources and tasks. The main contributions of this paper are:
\begin{itemize}
    \item On local self-attention, we show how to exploit a standard transformer model for long-span summarization, and we show good design considerations based on empirical results.
    \item On content selection, we demonstrate the best selection method at training time, and we propose a multitask content selection (MCS) method outperforming baselines at test time.
    \item Our work has set new state-of-the-art results on Spotify Podcast, arXiv and PubMed datasets in the ROUGE scores. Furthermore, with a small-scale GPU card, our approach achieves comparable or superior performance to previous state-of-the-art systems.
\end{itemize}

\begin{figure*}[!t]
\centering
\includegraphics[width=0.99\textwidth,keepaspectratio]{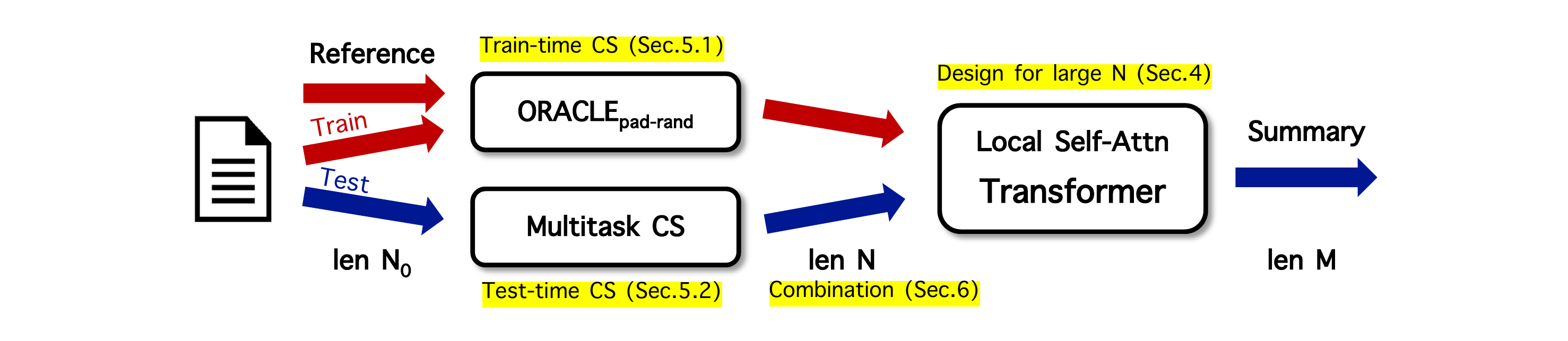}
\caption{Overview of the combined architecture where we highlight different aspects of this work. $N_0$ is the original document length, $N$ is the input length to the generation system, and $M$ is the summary length.}
\label{fig:architecture}
\end{figure*}

\vspace{-0.28cm}
\section{Related Work}
\vspace{-0.04cm}
\textbf{Efficient Transformers.}
Pre-trained transformer models have shown success and become the starting point for various NLP problems such as BERT \cite{devlin2018bert} in contextual representation, GPT2 in text generation \cite{radford2019language}, or BART in seq2seq tasks \cite{lewis-etal-2020-bart}. However, the memory and time requirements for transformer models grow quadratically with the sequence length, and for long-span tasks this quickly leads to GPU running out of memory in training. To mitigate the quadratic nature, a wide range of modified architectures have recently been proposed \cite{tay2021long}. They reduce the quadratic complexity of the full self-attention mechanism by using fixed attention patterns \cite{ parmar2018image, dai-etal-2019-transformer, child2019generating, qiu-etal-2020-blockwise, ainslie-etal-2020-etc, zaheer2020big, beltagy2020longformer}, learnable patterns \cite{kitaev2020reformer, tay2020sparse}, low-rank matrix approximation \cite{wang2020linformer}, or kernel method \cite{choromanski2021rethinking}. Alternatively, it has been shown that some attention heads are redundant and can be pruned to reduce model size \cite{voita-etal-2019-analyzing, michel_sixteen_heads}. Knowledge distillation reduces memory and compute by compressing a large model to a smaller one \cite{hinton2015distilling, sanh2019distilbert}. In contrast, we focus on the dependencies of long input and target sequences in encoder-decoder architectures, and we exploit publicly available transformer models with summarization weights to long-span summarization tasks.

\vspace{8pt}
\noindent \textbf{Long-span Summarization.} Efficient transformer architectures have been applied to summarize long documents such as BigBird \cite{zaheer2020big}, and Longformer-Encoder-Decoder (LED) \cite{beltagy2020longformer}, which has recently been revised parallel to this work.\footnote{On the self-attention aspect, we believe this system is the most comparable to ours (see comparisons in Sec. \ref{section:arxiv_results}).} Hierarchical transformer architectures have been applied to multi-document summarization \cite{liu-lapata-2019-hierarchical}, and extractive news and table-to-text summarization \cite{zhang-etal-2019-hibert, narayan-etal-2020-stepwise}. Hierarchical attention RNN system has been applied to summarize long articles \cite{cohan-etal-2018-discourse}.

Alternatively, earlier methods show that good content selection helps abstractive news summarization systems \cite{chen-bansal-2018-fast, gehrmann-etal-2018-bottom, hsu-etal-2018-unified}. Hybrid systems that select sentences and generate an abstractive summary have been proposed such as extractive system + TLM for scientific articles \cite{pilault-etal-2020-extractive}, simple selection + BART for podcasts \cite{manakul2020cued_speech, kaiqiang_trec2020}, and guided summarization by BERT-based keyword/sentence extraction + BART for news and scientific articles \cite{he2020ctrlsum, dou-etal-2021-gsum}.

Other work includes dividing the source and target into multiple smaller pairs to train abstractive summarizers \cite{gidiotis2020divide}. Extractive methods with and without redundancy reduction techniques for long-span summarization have been studied \cite{xiao-carenini-2019-extractive, xiao-carenini-2020-systematically}.

\section{Experimental Setup}
\subsection{Dataset}
\textbf{Spotify Podcast.}\footnote{\hspace{-1pt}\url{https://podcastsdataset.byspotify.com}} The dataset consists of ASR transcripts with human descriptions as summaries \cite{clifton-etal-2020-100000}. We follow the data processing at TREC2020 \cite{jones_trec2020} in removing bad transcript-summary pairs from a total of 105,360+1,027 episodes, resulting in train/valid/test splits of 60,415/2,189/1,027 episodes the same as \citet{manakul2020cued_speech}.

\vspace{4pt}
\noindent \textbf{arXiv and PubMed.} Popular long document summarization datasets consist of academic articles with abstracts as summaries \cite{cohan-etal-2018-discourse} and train/valid/test splits of 203,037/6,436/6,440 for arXiv and 119,924/6,633/6,658 for PubMed.

\begin{table}[ht]
  \centering
\scalebox{0.9}{
  \begin{tabular}{rcccc}
    \toprule
    Dataset &\#Doc  &Input &90$^{\text{th}}$\% &Target \\
    \midrule
    Podcast &106k  &5,727 &11,677 &61.1 \\
    arXiv   &216k  &8,584 &16,108  &367 \\
    PubMed  &133k  &3,865 &7,234   &260 \\
    \bottomrule
      \end{tabular}
}
  \caption{Length Statistics (mean \& 90$^{\text{th}}$\%-ile).}
  \label{tab:data_statistics}
\end{table}
\vspace{-2pt}
\subsection{Models}
\label{section:models}
\begin{figure}[!ht]
    \centering

    \begin{subfigure}[b]{0.50\linewidth}
    \centering
      \includegraphics[width=\linewidth,height=2.2cm,keepaspectratio]{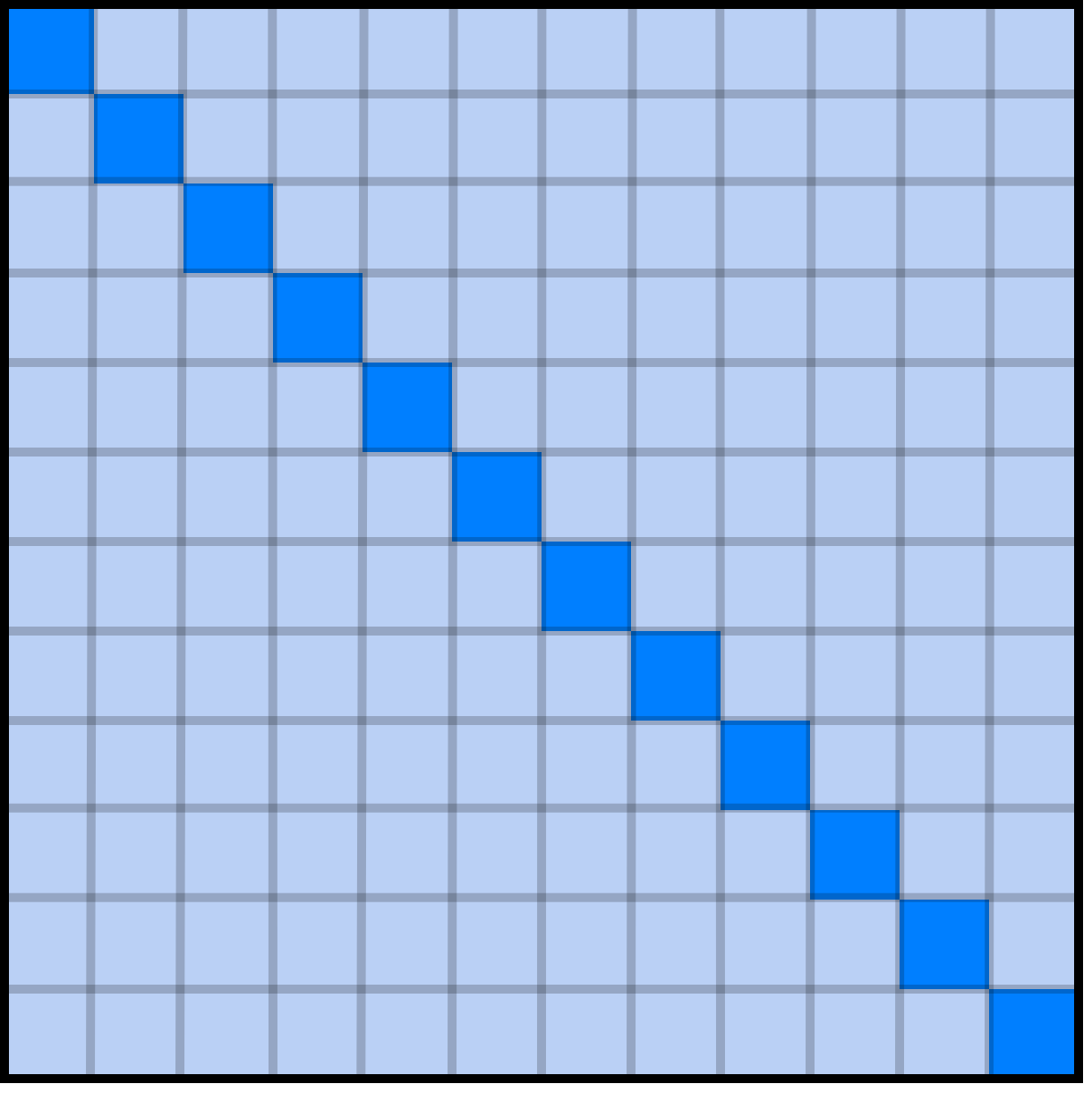}
    \caption{Full}
    \label{fig:full_attn}
    \end{subfigure}%
    \hfill
    \begin{subfigure}[b]{0.50\linewidth}
    \centering
      \includegraphics[width=\linewidth,height=2.2cm,keepaspectratio]{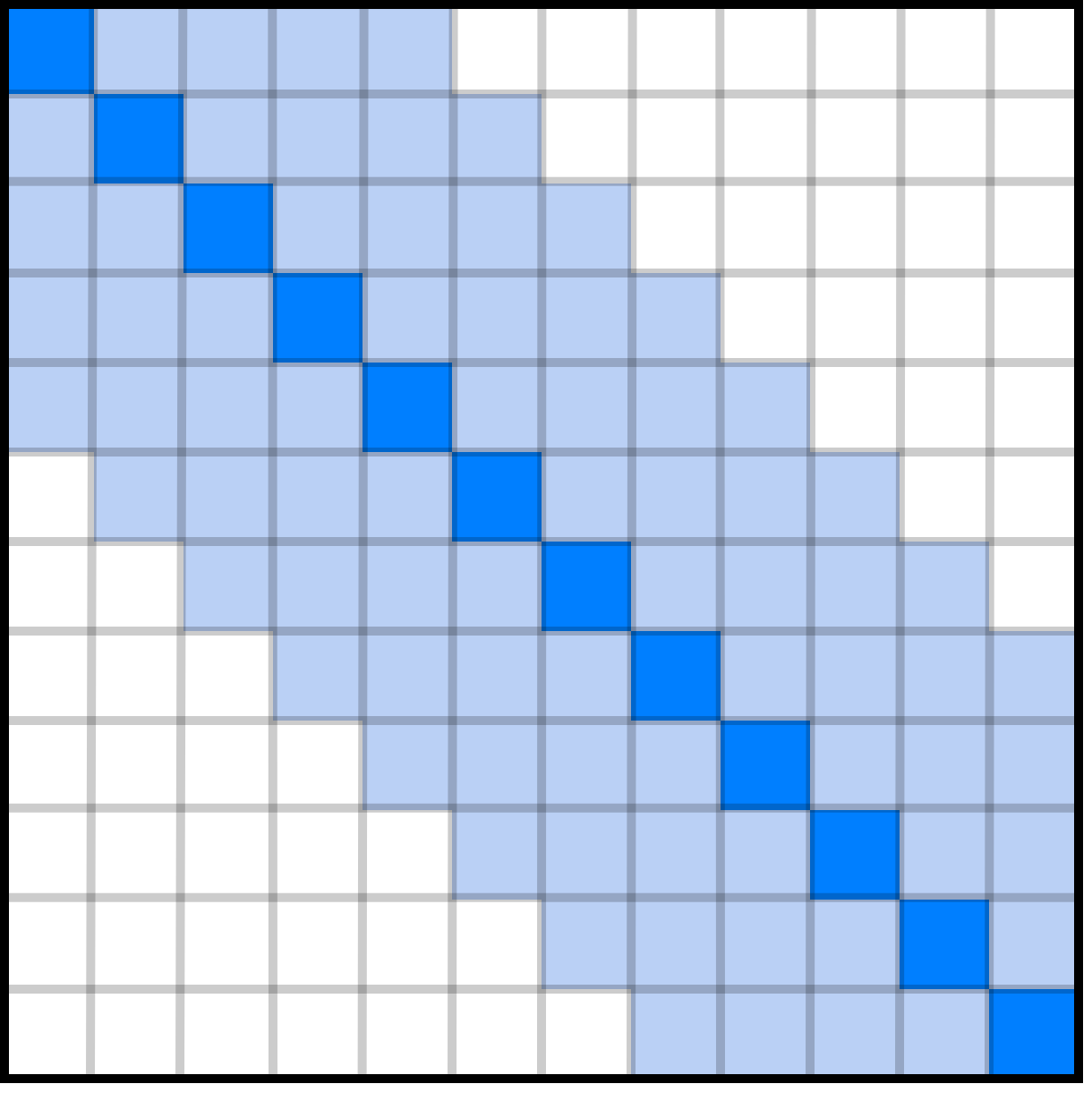}
    \caption{Local ($W$=9)}
    \label{fig:local_attn}
    \end{subfigure}%
    \caption{Self-Attention Pattern.}
    \label{fig:attn_mech}
\end{figure}

\noindent \textbf{BART and LoBART.} We use the publicly released BART model \cite{lewis-etal-2020-bart} fine-tuned on CNNDM \cite{hermann2015teaching}.\footnote{\hspace{-1pt}\url{https://huggingface.co/facebook/bart-large-cnn}} Following the local window attention in Sparse Transformer \cite{child2019generating} and Longformer \cite{beltagy2020longformer}, we modify the self-attention mechanism in the encoder to local self-attention (see Figure \ref{fig:attn_mech}), and we refer to this local self-attention BART as LoBART. It has the same architecture as BART, e.g. the number of parameters, except that we extend positional embedding beyond 1,024 by copying BART's positional embedding with flipping to allow a smoother transition. See details in Appendix \ref{appendix:model_parameters}.

\noindent \textbf{Hierarchical RNN.} The content selection model is based on a hierarchical encoder-decoder architecture that has been shown effective on meeting and long document summarization \cite{cohan-etal-2018-discourse, hier_rnn_2019, li-etal-2019-keep}. The model consists of word-level and sentence-level GRUs \cite{cho-etal-2014-learning}. We add a linear layer on top of the sentence-level GRU to perform extractive labelling. The sentence-level attention mechanism and extractive labelling modules form our multitask content selection (MCS). More details in Section \ref{section:mcs}.

We provide the full details about our implementation, model parameters, hyperparameters, optimizer, and training configurations in Appendix \ref{appendix:implementation_details}.


\section{Longer Span via Local Self-Attention}
\label{section:local_attention}

It has been known that memory and compute complexity of transformers is \textit{quadratic} with the sequence length. However, in encoder-decoder architectures, the exact dependencies on input length $N$, target length $M$, and batch size $B$ are less understood. This is particularly important in long-span seq2seq tasks because large memory or compute requirement could make training impractical. Thus, this work studies these dependencies, and shows the trade-off between the size of input span and the size of attention span in local self-attention.

\subsection{Memory Analysis and LoBART Design}





Firstly, through a regression analysis for an encoder-decoder architecture such as BART, the memory required in training is:
\begin{equation*}
    c^b_1 +  B(c^b_2 M + c^b_3 N + c^b_4 MN + c^b_5 M^2 + c^b_6 N^2)
\end{equation*}
The term $c^b_1$ depends on only the model size and optimizer, and it is \textit{constant} (theoretical calculation provided in Appendix \ref{appendix:complexity}). The remaining terms are activation memory associated with the activation outputs cached for backpropagation, and they grow with $N$, $M$, and $B$. Table \ref{tab:memory_profile} shows system-independent\footnote{system-independent across hardware and machines; albeit implementation-dependent. This analysis is based on widely used PyTorch and Huggingface implementation.} regression results for the memory in training BART. It is apparent that as $N$ grows the dominant term is $c^b_6N^2$, which is associated with the encoder self-attention. Thus, this motivates us to modify self-attention only on the encoder side.

\begin{table}[ht]
\tabcolsep=0.17cm
  \centering
\scalebox{0.9}{
  \begin{tabular}{c|cccccc}
    \toprule
    Term &$c^b_1$  &$c^b_2M$ &$c^b_3N$ &$c^b_4 MN$ &$c^b_5M^2$ &$c^b_6 N^2$\\
    \midrule
    GiB &6.05  &0.23 &0.84 &0.21 &0.02 &1.53 \\
    \bottomrule
  \end{tabular}}
  \caption{BART's Memory Profile ($N$=1024, $M$=144).}
  \label{tab:memory_profile}
\end{table}

\noindent By introducing local self-attention of width $W$, the memory in training LoBART becomes:
\begin{equation*}
     c^l_1 +  B(c^l_2 M + c^l_3 N + c^l_4 MN + c^l_5 M^2 + c^l_6 NW)
\end{equation*}
For large $N$, the memory is now dominated by $c^l_6 NW$. The coefficient $c^l_6\approx 1.72 c^b_6$, suggesting that $W$ should be at most $0.58N$ to reduce memory. We provide more details about the exact theoretical calculation for model and optimizer memory as well as time complexity in Appendix \ref{appendix:complexity}.

\begin{figure}[ht]
    \centering
      \includegraphics[width=0.80\linewidth,keepaspectratio]{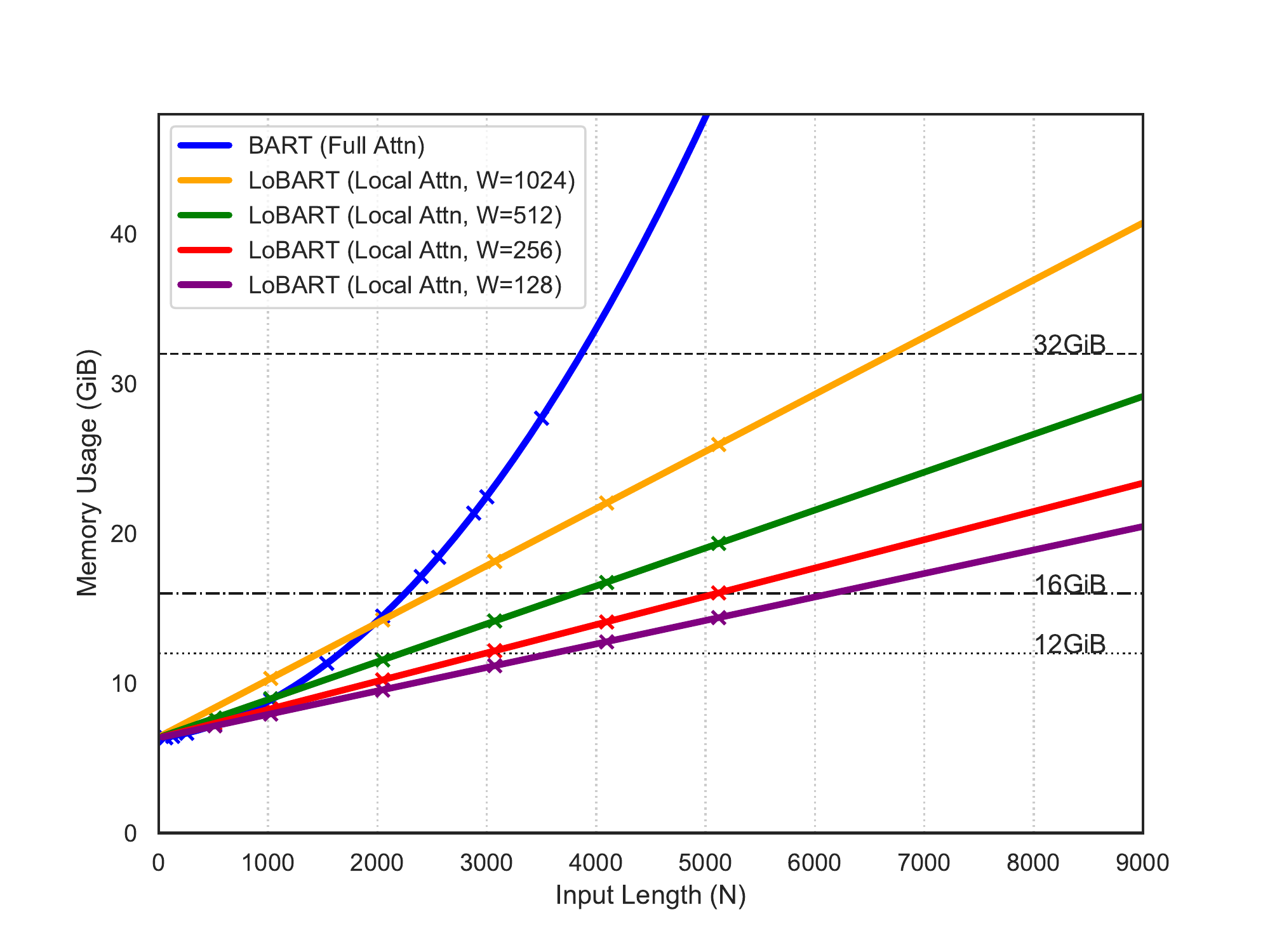}
    \caption{Operating points for $B$=1 and $M$=144. (1) Section \ref{section:local_attention} studies local attention to reduce quadratic complexity to linear. As $W$ decreases, the gradient of linear complexity decreases. (2) Section \ref{section:content_selection} studies content selection to move an operating point to the left.}
    \label{fig:memory_pred}
\end{figure}

The memory for training BART/LoBART in Figure \ref{fig:memory_pred} enables us to choose an operating point. Additionally, other \textit{complementary} techniques for reducing memory in training include: (i) gradient-checkpoint where a subset of intermediate values in the computation graph are cached, and the rest are re-computed during backpropagation \cite{chen2016training}, but this requires changes to optimization and leads to longer training time; (ii) half/mixed-precision training \cite{micikevicius2017mixed} that would almost halve y-axis in Figure \ref{fig:memory_pred}, but this requires changes to the model precision and may result in lower performance; (iii) model parallelism with micro-batching \cite{huang2019gpipe}, but this method requires multiple accelerators.

\subsection{BART and LoBART}
\begin{figure}[ht]
    \centering
      \includegraphics[width=0.88\linewidth,height=8cm,keepaspectratio]{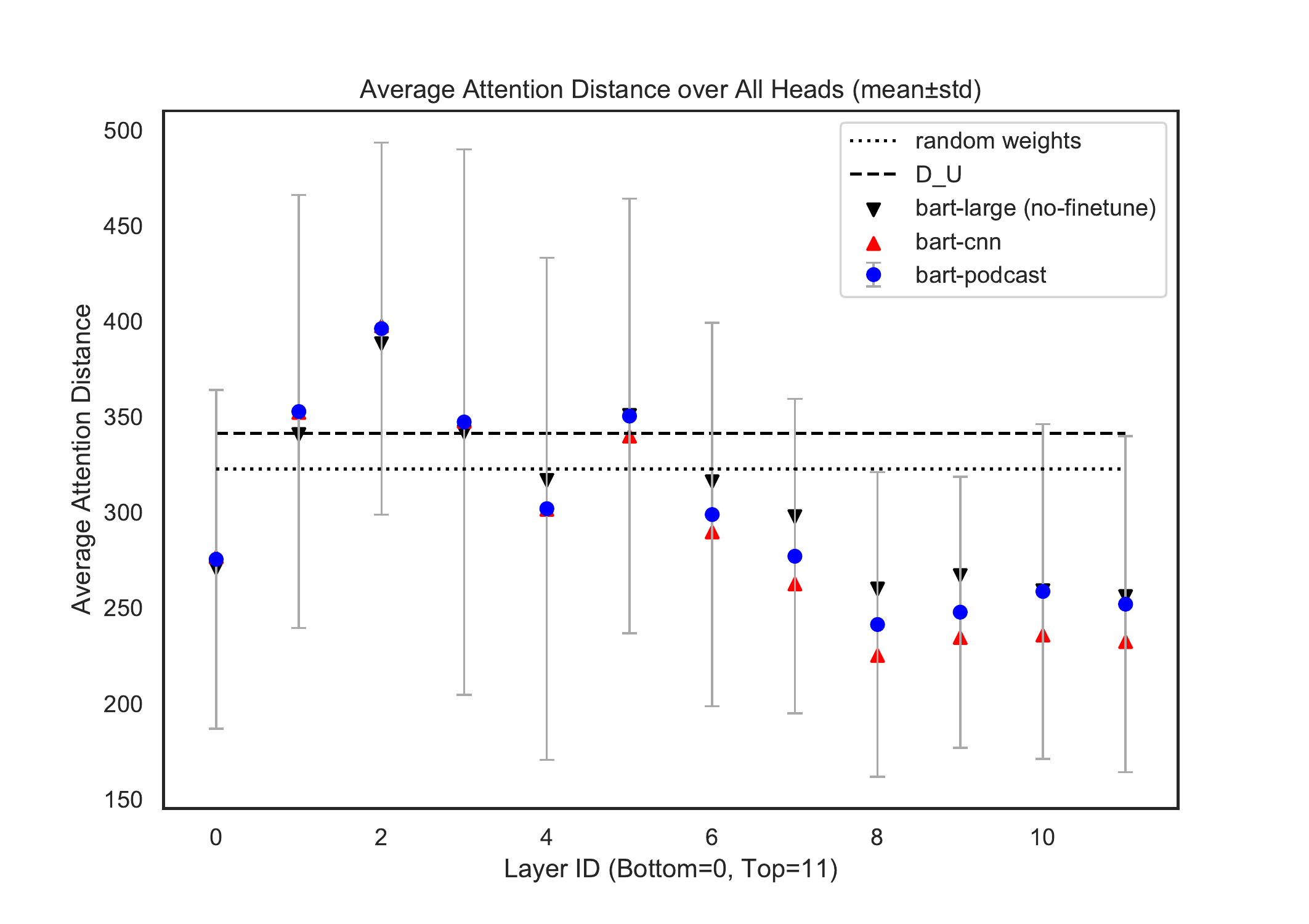}
    \caption{The average mean distance across multi-heads for each layer. The average mean distance of the random weight model is slightly lower than $D_U$ as some inputs are shorter than 1,024.}
    \label{fig:mean_distance}
\end{figure}

We study the characteristics of the full self-attention in BART by defining the mean attention distance in a particular layer and head as follows:

\begin{equation}
    D = \frac{1}{N} \sum_{i=1}^N \left( \sum_{j=1}^N \alpha_{i,j} \times |i-j| \right)
\end{equation}
where $\alpha_{i,j}$ is the attention weight of position $i$ attending to position $j$ ($\sum_{j=1}^N \alpha_{i,j} = 1$). This measure corresponds to the average distance of self-attention. If the attention weight is uniform, $D_{U} = \frac{N^2-1}{3N}$. For $N=1024$, $D_{U} = 341$. In Figure \ref{fig:mean_distance}, our results show that most layers have a shorter mean distance than $D_U$, supporting that the information is more localized. The mean distances of differently initialized BART models computed on the podcast data also show that the attention mechanism is learned during pre-training stage as there is little variation after the pre-training stage. As illustrated in Figure \ref{fig:mean_distance}, the average attention distance $D$ of the BART model is around 250-350 tokens. This suggests the window size $W$ should be designed to be above 700, allowing half local attention window $W/2$ be greater than 250-350 to effectively match BART and to exploit transfer learning more efficiently.

Subsequently, we train different configurations of BART/LoBART models up to our GPU memory limit of 32GiB. The results in Table \ref{tab:bart_1024_2048_4096_new} show that: (i) expanding the model to accommodate longer input spans improve over the baseline BART(1k) as opposed to \citet{manakul2020cued_speech} that trained longer-span models by freezing bottom layers and did not show any improvement over their baseline; (ii) Although LoBART(8k) with $W$=512 can process longer input spans than LoBART(4k) with $W$=1024, it performs worse and we suggest that this is because LoBART(8k)'s window is too small, e.g. $<$700, to utilize transfer learning efficiently and its effective receptive field is also smaller.

\begin{table}[ht]
\tabcolsep=0.20cm
  \centering
\scalebox{0.9}{
  \begin{tabular}{lc|c|ccc}
    \toprule
  System &$W$ &{GiB}  &R1 &R2 &RL \\
    \midrule
BART(1k) &Full  &8.9  &26.43 &9.22  &18.35  \\ 
    \midrule
LoBART(2k) &128   &9.6   &25.88 &8.89 &17.87 \\ 
LoBART(2k) &256   &10.2  &25.93 &8.80 &17.82 \\ 
LoBART(2k) &512   &11.6  &26.35 &8.98 &18.19  \\ 
LoBART(2k) &1024  &14.2  &26.44 &9.26 &18.25 \\ 
BART(2k) &Full  &14.5  &26.63 &9.41 &18.65 \\ 
    \midrule
LoBART(4k) &128   &12.8   &26.42 &9.02 &18.12 \\ 
LoBART(4k) &256   &14.1   &26.66 &9.22 &18.33 \\ 
LoBART(4k) &512  &16.7    &26.75 &9.54 &18.54 \\ 
LoBART(4k) &1024  &22.0   &\textbf{27.02} &\textbf{9.57} &\textbf{18.78} \\ 
    \midrule
LoBART(8k) &128  &19.3 &26.45 &9.04 &18.23 \\ %
LoBART(8k) &256  &21.1 &26.72 &9.30 &18.36 \\ 
LoBART(8k) &512  &27.1 &26.90 &9.47 &18.50 \\ 
    \bottomrule
  \end{tabular}
}
  \caption{BART \& LoBART memory requirement in training and performance. ($n$k) denotes maximum input length of $n\times1024$.}
  \label{tab:bart_1024_2048_4096_new}
\end{table}

\section{Longer Span via Content Selection}
\label{section:content_selection}
Some input sequences still exceed LoBART's longer fixed-span limit. Further extending the input span would lead to a small local attention span, a diminishing improvement, or GPU running out of memory. Alternatively, it has been shown that a better content selection improves abstractive summarization in news \cite{chen-bansal-2018-fast, gehrmann-etal-2018-bottom, hsu-etal-2018-unified}, multi documents \cite{liu-lapata-2019-hierarchical, liu2018generating}, and scientific articles \cite{pilault-etal-2020-extractive}. Thus, we propose to tackle the excess length by content selection. Here, we distinguish between two phases of content selection: training time and test time.

\subsection{Training-time Content Selection}
\label{section:training_bart_with_cs}
During training, ground-truth targets are available. We categorize selection methods in this phase into two types: ground-truth based (model-free), which is also referred to as \textit{oracle}; and model-based. Ground-truth based methods cannot be used at test time, while model-based methods can be applied at both phases. Although model-based methods do not rely on ground-truth targets, they have the advantage of matching in training and test phases. Existing oracle methods include using ROUGE-2 recall \cite{liu2018generating} or the average of ROUGE-1,2,L recall \cite{pilault-etal-2020-extractive}. We discuss model-based methods in Section \ref{section:mcs}, where we propose the MCS method. Let the subscript $(i,j)$ denote the position of the $j$-th word in the $i$-th input sentence, the full input $ \mathbf{X}=\{\mathbf{x}_1,...,\mathbf{x}_i,...,\mathbf{x}_{N_1}\} = [\underbrace{x_{1,1},{x}_{1,2},{x}_{1,J_1}}_{\text{sent }1},...,\underbrace{{x}_{i,1},{x}_{i,J_i}}_{\text{sent }i},...,\underbrace{{x}_{N_1,1},{x}_{N_1,J_{N_1}}}_{\text{sent }N_1}]$.
Content selection re-ranks, truncates, and sorts $\mathbf{X}$ to get ${\mathbf{X}}^{\text{cs}}$ for training BART/LoBART as follows:
\begin{align}
    \bar{\mathbf{X}} &= \{\mathbf{x}_{r_1},\mathbf{x}_{r_2},\mathbf{x}_{r_3},...,\mathbf{x}_{r_R}\} \\
    {\mathbf{X}}^{\text{cs}} &= {\tt SortOrig}({\tt TruncateN}(\bar{\mathbf{X}}))
\end{align}
where $r_i$ is the index of the sentence of rank $i$, the ${\tt TruncateN}$ operation filters $\bar{\mathbf{X}}$ such that the total of number of words is less than $N$, and ${\tt SortOrig}$ retains the original sentence order. The following ranking methods are considered:
\begin{itemize}
    \item Truncation (TRC): $r_k = k$.
    \item Model-based: Given the score $f$ of model $\boldsymbol{\phi}$, \\ $r_k = \{i \in N_1 : f_{\boldsymbol{\phi}}(i|\mathbf{X}) \hspace{4pt} \text{is ranked} \hspace{4pt} k\text{-th}\}$
    \item Oracle (ORC): Given the ground-truth summary $\mathbf{y}$ and similarity measure $d$, \\ $r_k = \{i \in N_1 : d(\mathbf{x}_i, \mathbf{y}) \hspace{4pt} \text{is ranked} \hspace{4pt} k\text{-th}\}$
\end{itemize}
In this work, we use ROUGE-2 recall as the similarity measure $d$. For the ORC method, first, we retain only sentences with positive $d$, leading to $R \leq N_1$. We found that the number of sentences with positive $d$ is low at 21.3\% of the total number of sentences in average on podcast data. This corresponds to 56\% of training instances being shorter than BART input span of 1024.\footnote{We refer to this percentage as \%AgORC$_\text{no-pad}$ (the percentage of inputs aggressively extracted by the oracle method).} This no-padding oracle method ({ORC}$_{\text{no-pad}}$) is highly \textit{aggressive}, potentially preventing the downstream summarizer from learning complex abstraction. Hence, we propose variants of oracle methods to extend the {ORC}$_{\text{no-pad}}$-selected input to the max input span $N$:
\begin{itemize}
    \item {ORC}$_{\text{pad-lead}}$: Pad by leading unselected sentences and keep the original sentence order.
    \item {ORC}$_{\text{pad-rand}}$: Pad by random unselected sentences and keep the original sentence order.
\end{itemize}

\begin{figure}[ht]
    \centering
      \includegraphics[width=0.95\linewidth,height=8cm,keepaspectratio]{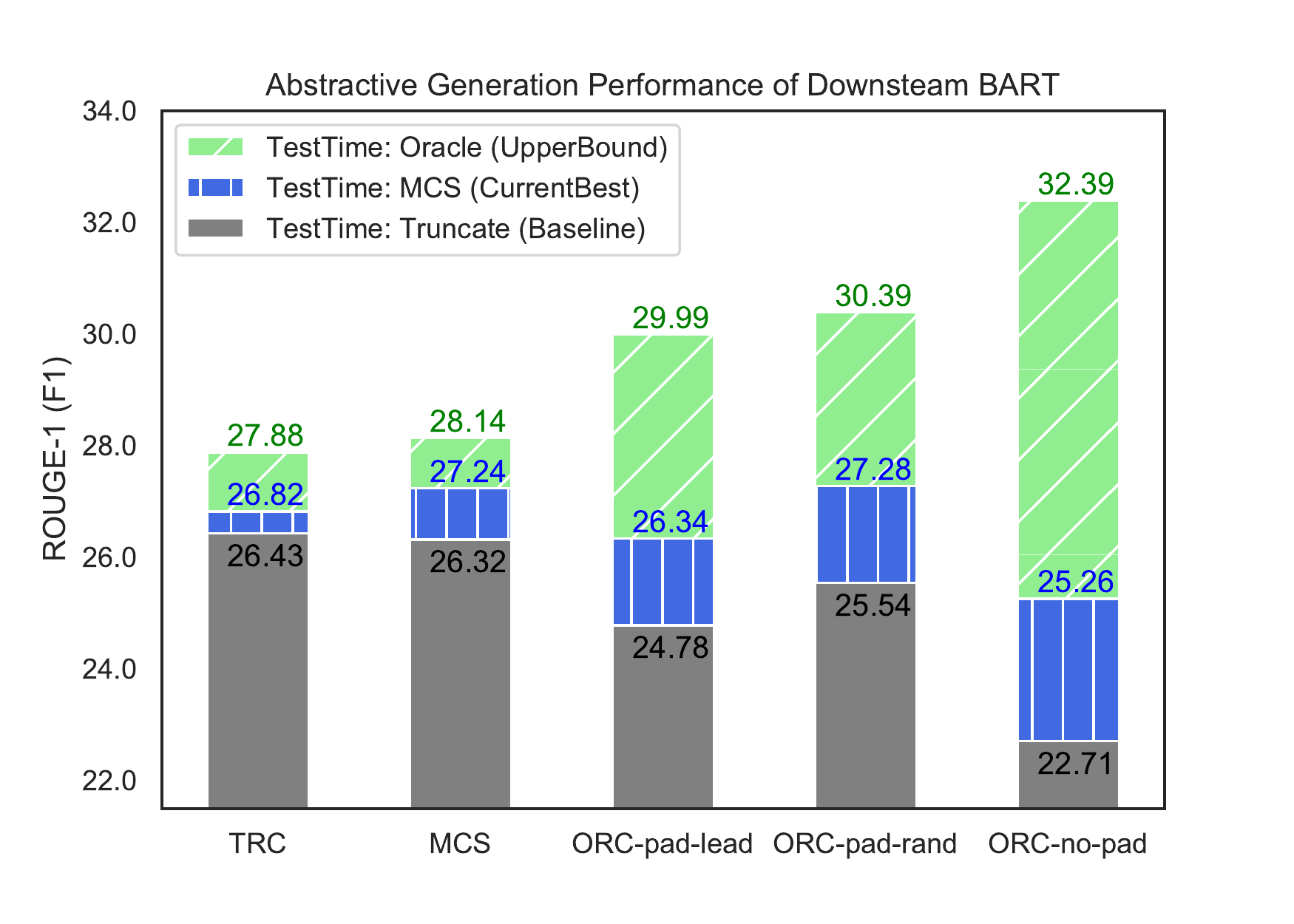}
    \caption{The impact of training-time content selection methods on BART(1k) performance.}
    \label{fig:barchart}
\end{figure}

In Figure \ref{fig:barchart}, since any oracle method is considered cheating at test time, the best performance is obtained by MCS (in blue), and the upper bound performance is obtained by optimal oracle method (in green). The results show that although {ORC}$_{\text{no-pad}}$ yields the highest upper bound, the abstractive model in fact does not learn how to perform abstraction. For instance, with TRC or MCS at test time, {ORC}$_{\text{no-pad}}$ yields the lowest performance level.  The best way to fine-tune the abstractive model shown in Figure \ref{fig:barchart} is using {ORC}$_{\text{pad-rand}}$. Compared to {ORC}$_{\text{pad-lead}}$, {ORC}$_{\text{pad-rand}}$ is better as it introduces more diversity to the abstractive model. Compared to the model-based method, {ORC}$_{\text{pad-rand}}$ is also computationally less expensive.

In addition, Table \ref{tab:combine_podcast} shows that when there is no content selection at test time (i.e. TRC applied), LoBART(4k) and LoBART(8k) benefit from {ORC}$_{\text{pad-rand}}$, whereas BART(1k) does not. This is because in the 1k setting, content selection is more aggressive; as a result, the large mismatch between training and test leads to a poor result. Thus, we suggest that the best content selection during training is {ORC}$_{\text{pad-rand}}$ given that content selection will be used at test time, or model's input span is long.

\subsection{Multitask Content Selection (MCS)}
\label{section:mcs}
To process long input sequences entirely, we consider RNN, whose memory requirement grows linearly with the sequence length, and hierarchical architectures which have been shown effective for long seq2seq tasks \cite{cohan-etal-2018-discourse, li-etal-2019-keep}. In this work, the hierarchical RNN model described in Section \ref{section:models} has memory requirement given the target length of 144 during training of $0.83 + B(3.96\times10^{-5}+3.33\times10^{-5}N_2)N_1$,\footnote{Obtained by least-squares regression with 20 samples.} where $N_1$ is \#sentences, and $N_2$ is the maximum number of words in a sentence, and $B$ is batch size. By setting $N_1$=1000 and $N_2$=50, only 2\% of podcast data exceeds this limit, while taking GPU memory to only 2.53GiB for $B$=1. Thus, this shows that this model can cover long sequences.

Previous model-based methods treat content selection as extractive labelling and create labels heuristically \cite{pilault-etal-2020-extractive}, or using encoder-decoder attention mechanism \cite{manakul2020cued_speech}. To utilize both of these in one framework, we propose a Multitask Content Selection (MCS) method where we train the hierarchical encoder-decoder with attention mechanism and a classification layer on top of the encoder (described in Section \ref{section:models}). First, the model is trained on seq2seq abstractive summarization objective:
\begin{equation}
    \mathcal{L}_{\text{seq2seq}} = -\sum_{m=1}^M \log P(y_m | \mathbf{y}_{<m},\mathbf{X})
\end{equation}
Second, we create binary labels as follows: for sentence $i$, the label $z_i$ is 1 if $d(\mathbf{x}_i,\mathbf{y}) > 0$; else $z_i$ is 0, and $d$ is the ROUGE-2 recall measure. The extractive labelling task objective is:
\begin{equation}
\resizebox{.89\hsize}{!}{$
    \mathcal{L}_{\text{label}} = -\sum_{i=1}^{N_1} \left( z_i \log \hat{z}_i + (1-z_i) \log (1 - \hat{z}_i) \right)
$}
\end{equation}
\vspace{-0.5cm}
\begin{equation}
    \hat{z}_i =\text{sigmoid} (\mathbf{W}_{\text{cls}} ^{T} \mathbf{h}_i + \mathbf{b}_{\text{cls}})
\end{equation}
where $\mathbf{h}_i$ is the sentence-level encoder output associated with sentence $i$, and $\mathbf{W}_{\text{cls}}, \mathbf{b}_{\text{cls}}$ are the parameters of the classification layer. Thus, the MCS training loss is defined as follows:
\begin{equation}
    \mathcal{L}_{\text{MCS}} = \gamma\mathcal{L}_{\text{label}} + (1-\gamma)\mathcal{L}_{\text{seq2seq}}
\end{equation}
At inference stage, there are two modes: (i) standard abstractive summary generation, e.g. via beam search decoding; (ii) ranking input sentences via labelling score and seq2seq attention score. The latter is how we use MCS during inference.\footnote{In practice, we run beam search decoding of width 4, and we obtain the attention score from the top beam.} For sentence $i$, the scores are:
\begin{equation}
\resizebox{.95\hsize}{!}{$
 \text{score}_{i,(\text{label})} = \hat{z}_i, \hspace{0.1cm} \text{score}_{i,(\text{seq2seq})} = \sum_{m=1}^M \alpha^{\tt s}_{m,i}
 $}
 \label{eq:mcs_inference_score}
\end{equation}
where $\alpha^{\tt s}_{m,i}$ is the sentence-level attention weight at decoder step $m$ over input sentence $i$. Since the scores are on different scales, rather than using the scores defined in Eq. \ref{eq:mcs_inference_score}, we simply rank the scores, and then normalize the score ranks into the range 0.0 to 1.0. Let nscore denote the normalized ranking score, the MCS inference score is:
\begin{equation}
   f_{\boldsymbol{\phi}}(i|\mathbf{X})  = \text{nscore}_{i,(\text{label})} + \text{nscore}_{i,(\text{seq2seq})}
\end{equation}
In our preliminary experiments, we vary the amount of selected sentences from the limit of BART/LoBART to a few sentences, and we found that more aggressive selection at test time degrades the performance. Therefore, our MCS selects input sentences up to the limit of BART/LoBART.

By setting $\gamma$=0.0, our method is comparable to the attention-based method in \citet{manakul2020cued_speech}. By setting $\gamma$=1.0, our method is similar to the extractive models in \citet{hsu-etal-2018-unified, pilault-etal-2020-extractive}.
In Table \ref{tab:mcs_results}, we show that when coupled with BART, MCS yields better summarization performance than both Attn-only and Ext-only baselines. MCS also achieves higher recall rate of sentences with $d(\mathbf{x}_i,\mathbf{y})>0$ than the two baselines.

\begin{table}[ht]
  \centering
\scalebox{0.9}{
  \begin{tabular}{l|c|ccc}
    \toprule
   {System}  &{\%Recall}  &R1 &R2 &RL  \\
    \midrule
   Attn ($\mathcal{L}_{\text{{seq2seq}}}$)        &38.85   &26.90 &9.70  &18.78 \\ 
   Ext  ($\mathcal{L}_{\text{{label}}}$)       &35.26    &26.39 &8.90 &18.03 \\ 
   \midrule
   MCS ($\mathcal{L}_{\text{{MCS}}}$)        &40.50 &27.28 &9.82  &19.00 \\  
    \bottomrule
  \end{tabular}
}
  \caption{The impact of test-time content selection methods on BART(1k) trained using ORC$_\text{pad-rand}$. Optimal $\gamma$=0.2 is tuned between 0.0-1.0 on the validation set.}
  \label{tab:mcs_results}
\end{table}

\section{Combined Approach}

\subsection{Spotify Podcast results}
In Table \ref{tab:combine_podcast}, a performance gain is obtained in all settings by adding MCS. By comparing different configurations with MCS, it can be seen that the gain from MCS in LoBART(8k) system is the lowest. This is because the average length is 5,727, meaning that many Podcasts inputs to LoBART(8k) do not benefit from content selection.

CUED-filt, the best single-model system in \citet{manakul2020cued_speech}, uses an attention-based content selection at both training and test time, and it is combined with fine-tuned vanilla BART. Our approach outperforms CUED-filt by improved content selection at both training time and test time as demonstrated by BART(1k)-ORC+MCS. Additionally, local self-attention allows training on longer sequences, and our LoBART(4k)-ORC+MCS system has yielded the best results. Lastly, even though LoBART(8k) requires more resource to train, it does not perform as well as LoBART(4k) due to its smaller attention window, and it also has a lower improvement when adding MCS.

\begin{table}[h]
\tabcolsep=0.14cm
  \centering
\scalebox{0.9}{
  \begin{tabular}{lcc|ccc}
    \toprule
 {System} &CS-trn &CS-tst &R1 &R2 &RL \\
    \midrule
CUED-filt$^*$ &\cmark &\cmark  &26.96 &9.75 &18.90 \\
   \midrule
BART(1k)   &\xmark &\xmark  &26.43 &9.22  &18.35  \\
BART(1k)   &\xmark &MCS  &26.82 &9.39 &18.57  \\
BART(1k)   &ORC    &\xmark  &25.54 &9.00 &17.83  \\ 
BART(1k)   &ORC    &MCS  &27.28 &9.82 &19.00 \\  
\midrule
LoBART(4k) &\xmark &\xmark &27.02 &9.57  &18.78  \\ 
LoBART(4k) &\xmark &MCS    &27.53 &9.95  &19.08  \\
LoBART(4k) &ORC    &\xmark &27.36 &10.04  &19.33  \\ 
LoBART(4k) &ORC    &MCS &\textbf{27.81} &\textbf{10.30}  &\textbf{19.61}   \\ 
\midrule
LoBART(8k) &\xmark &\xmark &26.90 &9.47  &18.50  \\ 
LoBART(8k) &\xmark &MCS    &27.02 &9.52  &18.62   \\
LoBART(8k) &ORC    &\xmark &27.16  &9.84  &19.08  \\ 
LoBART(8k) &ORC    &MCS &27.49 &9.98  &19.25  \\ 
    \bottomrule
  \end{tabular}
}
  \caption{Podcast Results. The impact of training-time ORC$_\text{pad-rand}$ and test-time MCS. $^*$CUED systems were the top systems by human evaluation at Spotify Challenge 2020; CUED systems use BART with a model-based (trained on $\mathcal{L}_{\text{seq2seq}}$) content selection in both training and test stages.}
  \label{tab:combine_podcast}
\end{table}

\begin{table*}[!t]
  \centering
\scalebox{0.9}{
  \begin{tabular}{ccl|ccc|ccc}
    \toprule
     &\multirow{2}{*}{Type}          &\multirow{2}{*}{System}         &\multicolumn{3}{c}{arXiv}   &\multicolumn{3}{c}{PubMed} \\
      &  &                &R1 &R2 &RL    &R1 &R2 &RL \\
    \midrule
    \parbox[t]{3pt}{\multirow{9}{*}{\rotatebox[origin=c]{90}{\small Previous Work}}} &Abs&Discourse-Aware \cite{cohan-etal-2018-discourse}   &35.80 &11.05 &31.80 &38.93 &15.37  &35.21 \\
    &Mix&Ext+TLM \cite{pilault-etal-2020-extractive}        &41.62 &14.69 &38.03      &42.13 &16.27 &39.21 \\
    &Ext&ExtSum-LG+Rd\cite{xiao-carenini-2020-systematically} &44.01	&17.79	&39.09 &45.30	&20.42	&40.95\\
    &Abs&Pegasus \cite{zhang2020pegasus}             &44.21&16.95&38.83 &45.97&20.15&41.34 \\ 
    &Abs&DANCER       \cite{gidiotis2020divide}     &45.01 &17.60 &40.56      &46.34 &19.97 &42.42 \\
    &Abs&BigBird(3k) \cite{zaheer2020big}          &46.63 &19.02 &41.77      &46.32 &20.65 &42.33 \\
    &Abs&LED(4k)     \cite{beltagy2020longformer}  &44.40 &17.94 &39.76      &- &- &- \\
    &Abs&LED(16k)    \cite{beltagy2020longformer}  &46.63 &19.62 &41.83      &- &- &- \\
    &Mix&CTRLsum(BART+BERT) \cite{he2020ctrlsum}  &46.91 &18.02 &42.14 &- &- &- \\
    \midrule
    \parbox[t]{3pt}{\multirow{4}{*}{\rotatebox[origin=c]{90}{\small This Work}}}&Abs&$^\dagger$BART(1k)     &44.96 &17.25 &39.76    &45.06 &18.27 &40.84 \\
    &Mix&$^\ddagger$BART(1k)+MCS &47.68 &19.77 &42.25    &46.49 &19.45 &42.04 \\
     &Abs&$^\ddagger$LoBART(4k)     &46.59 &18.72 &41.24  &47.47 &20.47 &43.02 \\
    &Mix&$^\ddagger$LoBART(4k)+MCS &\textbf{48.79} &\textbf{20.55} &\textbf{43.31}  &\textbf{48.06} &\textbf{20.96} &\textbf{43.56} \\

    \bottomrule
      \end{tabular}
}
  \caption{Results on arXiv and PubMed. $^\dagger$denotes TRC applied, and $^\ddagger$denotes ORC$_\text{pad-rand}$ applied at training time.}
  \label{tab:arxiv_pubmed_result}
\end{table*}

\subsection{ArXiv and PubMed results}
To verify the effectiveness of our systems, we re-train BART(1k) and LoBART(4k) on arXiv and PubMed datasets.
\label{section:arxiv_results}
Our training is different from Ext+TLM \cite{pilault-etal-2020-extractive} where their abstractive models are trained using inputs extracted from top two sentences in ROUGE recall for each target sentence without padding, similar to ORC$_\text{no-pad}$. Although in 1k setting, ORC$_\text{no-pad}$ yields \%AgORC$_\text{no-pad}$ (defined in Section \ref{section:training_bart_with_cs}) of only 2.8\% on arXiv (12\% on PubMed), in 4k setting this is 39\% on arXiv (71\% on PubMed). Based on the best configurations on podcast data, we train BART(1k) and LoBART(4k) using TRC or ORC$_\text{pad-rand}$ content selection, and we train the hierarchical model on arXiv/PubMed for MCS.

\vspace{6pt}
\noindent \textbf{ArXiv.} In Table \ref{tab:arxiv_pubmed_result}, both BART(1k)+MCS and LoBART(4k)+MCS outperform all existing systems. To better understand the advantages of our approach, the following systems are compared: CTRLsum versus our BART(1k) baseline; LED and BigBird versus our LoBART(4k) system.

CTRLsum extends BART by conditioning it with extracted keywords $\mathbf{v}$ using a BERT-based model, e.g. $p(\mathbf{y}|\mathbf{X},\mathbf{v})$. Their BERT-based model uses sliding window allowing it to extract $\mathbf{v}$ in long sequences, but their BART is still limited to the first 1,024 tokens. As a result, it performs better than BART(1k), but worse than BART(1k)+MCS.

LoBART(4k) has a similar architecture to LED(4k) without the global attention pattern for special tokens. Instead, our LoBART(4k) benefits from knowledge transferred from CNNDM and the ORC$_\text{pad-rand}$ training-time content selection, which yields a larger gain when MCS is applied, i.e. the system trained with truncated data has a smaller gain when MCS is applied. Transfer learning comparison and additional results on the impact of ORC$_\text{pad-rand}$ are provided in Appendix \ref{appendix:additional_results}.

Compared to BigBird, LoBART(4k) has a longer input span, e.g. 3,072 vs. 4,096. However, BigBird benefits from utilizing more recent summarization specific pre-training Pegasus \cite{zhang2020pegasus} which is better than our transfer learning. BigBird incorporates a global attention pattern similar to LED, and it also has a random attention pattern. Hence, LoBART without MCS performs worse.

Ultimately, we show that adding MCS to either BART(1k) or LoBART(4k) yields a significant improvement, resulting in state-of-the-art results in both settings. Moreover, although the gain from adding MCS is comparable to the gain observed in extending LED(4k) to LED(16k), the content selection method adds less training cost.

\vspace{6pt}
\noindent \textbf{PubMed.} Similarly, LoBART(4k)+MCS achieves state-of-the-art results shown in Table \ref{tab:arxiv_pubmed_result}. In contrast to the arXiv results, BART(1k)+MCS does not outperform LoBART(4k) nor BigBird, and the gain from MCS is not as high in both 1k and 4k settings.

\subsection{Local Attention v.s. MCS.}
Local attention yields better performance on PubMed, while MCS yields better performance on arXiv. To understand this discrepancy, a fine-grained analysis is conducted.
\begin{figure}[!ht]
    \centering
    \begin{subfigure}[b]{0.85\linewidth}
    \centering
      \includegraphics[width=\linewidth,height=8cm,keepaspectratio]{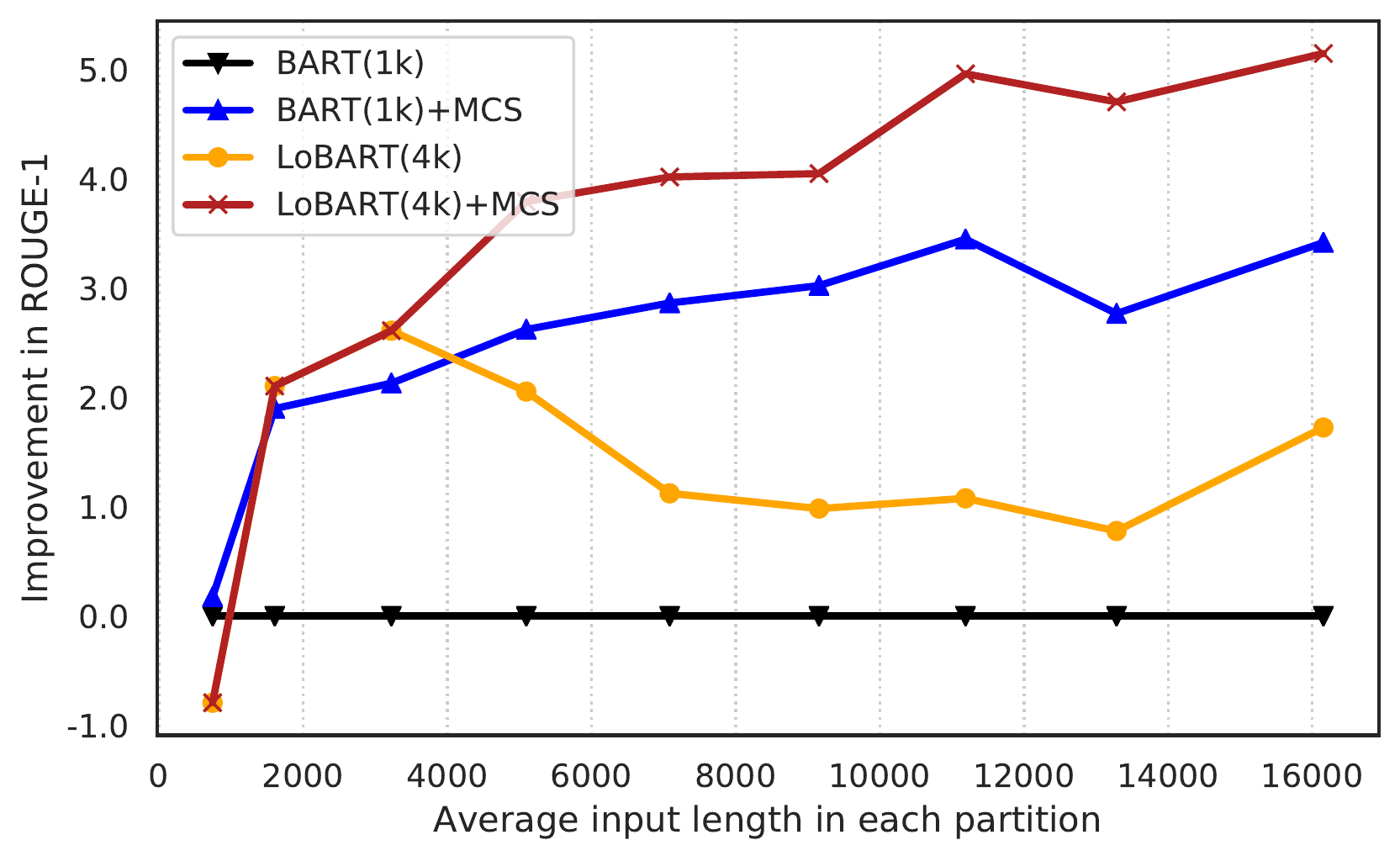}
    \caption{arXiv (Len:Avg=8,584, 90$^{\text{th}}$\%=16,108)}
    \label{fig:ablation_len_arxiv}
    \end{subfigure}%
    \vspace{8pt}
        \hfill
    \begin{subfigure}[b]{0.85\linewidth}
    \centering
      \includegraphics[width=\linewidth,height=8cm,keepaspectratio]{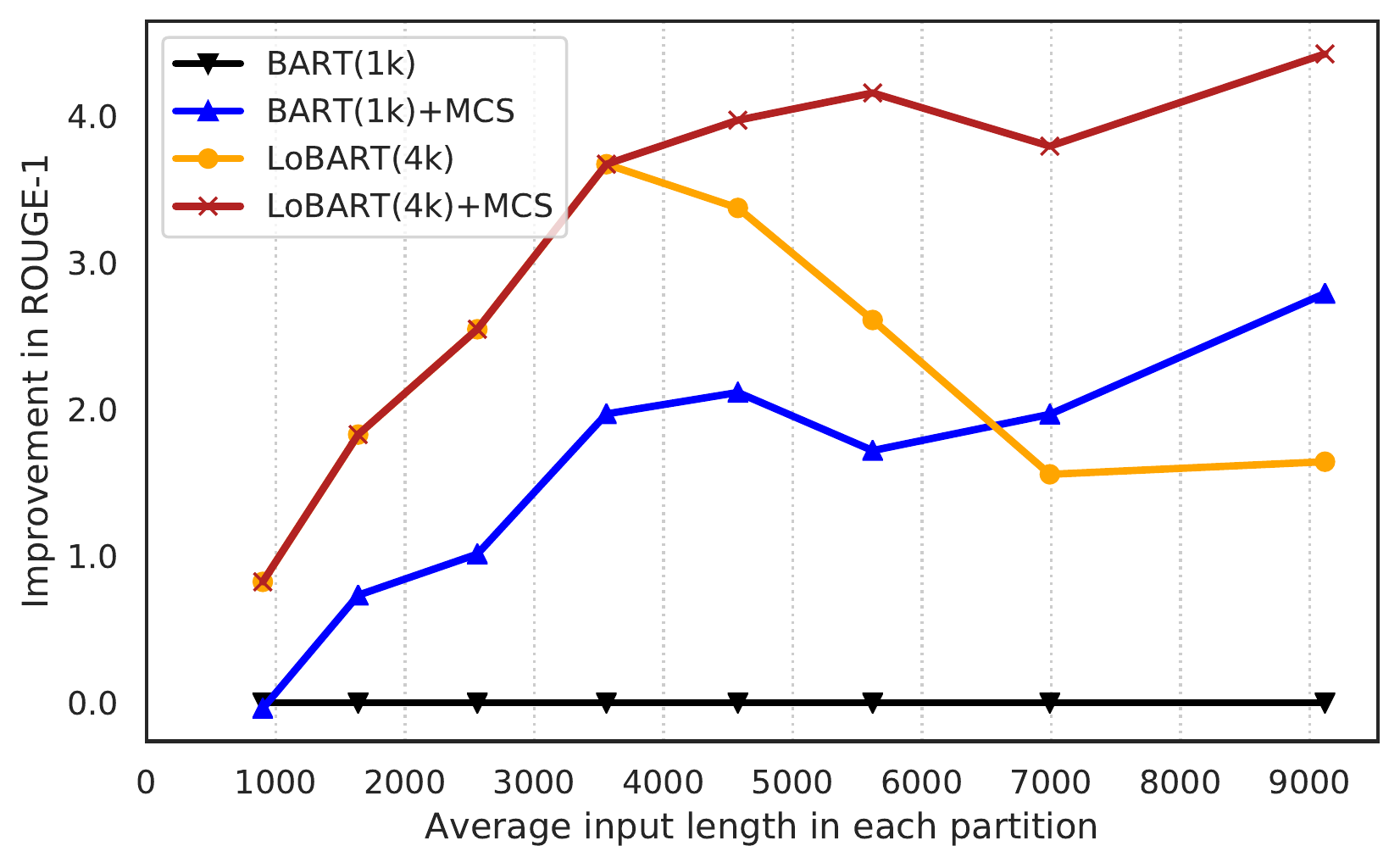}
    \caption{PubMed (Len:Avg=3,865, 90$^{\text{th}}$\%=7,234)}
    \label{fig:ablation_len_pubmed}
    \end{subfigure}%
    \caption{ROUGE-1 score relative to that of BART(1k) system evaluated on different partitions by length.}
    \label{fig:ablation_len}
\end{figure}

\noindent In Figure \ref{fig:ablation_len}, we partition the test sets by input lengths, and we evaluate the performance improvement in each partition with respect to the BART(1k) baseline.\footnote{For arXiv/PubMed, each test set consists of over 6,000 instances, while Podcast test set has only 1,027 instances. The same analysis is conducted on Podcast, but the results are noisy due to the smaller size of its test set (see Appendix \ref{appendix:additional_results}).} The results illustrate that as the input length $N$ increases:
\begin{itemize}
    \item The improvement of systems \textit{with} MCS increases and subsequently plateaus out.
    \item The improvement of systems \textit{without} MCS decreases once the input exceeds the length limit but then plateaus, suggesting that fixed-span systems without content selection perform worse once the maximum fixed-span is reached. For instance, below 4,000 input words, LoBART(4k) without MCS performs better than BART(1k)+MCS on both datasets.
\end{itemize}
Therefore, our MCS method is more effective on arXiv compared to PubMed because the average length of PubMed documents is more than twice shorter than the average length of arXiv documents.

\section{Conclusion}
We study two methods for long-span summarization tasks. First, on local self-attention transformers, we present the design considerations for local self-attention BART, and we investigate the feasibility and performance of different network configurations. Second, on content selection, we distinguish between training time and test time methods, and we provide a good practice for both phases. At training time, we show that the oracle method with random sentences padded (ORC$_\text{pad-rand}$) yields the best results. At test time, we propose multitask content selection (MCS) that shows an improvement over baselines. We demonstrate that content selection is essential, in particular for longer documents such as the articles in the arXiv dataset. Our BART(1k)+MCS outperforms the current best systems on Podcast and arXiv datasets, and this system does not require a large-scale accelerator in training. Ultimately, by combining local self-attention technique with MCS, our LoBART(4k)+MCS system has set new state-of-the-art results in terms of ROUGE scores in all three long-span summarization tasks. Future work will focus on training our LoBART+MCS system in an end-to-end fashion.

\section*{Acknowledgments}
This paper reports on research supported by ALTA institute, Cambridge Assessment English, University of Cambridge, and Cambridge International \& St John’s College Scholarship. Thanks to Yiting Lu, Qingyun Dou, Xixin Wu, Raf Czlonka, and Kate Knill for interesting discussions and computing resource support. Thanks to the anonymous reviewers for their helpful comments.
\bibliographystyle{acl_natbib}
\bibliography{anthology,acl2021}

\begin{thebibliography}{58}
\expandafter\ifx\csname natexlab\endcsname\relax\def\natexlab#1{#1}\fi

\bibitem[{Ainslie et~al.(2020)Ainslie, Ontanon, Alberti, Cvicek, Fisher, Pham,
  Ravula, Sanghai, Wang, and Yang}]{ainslie-etal-2020-etc}
Joshua Ainslie, Santiago Ontanon, Chris Alberti, Vaclav Cvicek, Zachary Fisher,
  Philip Pham, Anirudh Ravula, Sumit Sanghai, Qifan Wang, and Li~Yang. 2020.
\newblock \href {https://doi.org/10.18653/v1/2020.emnlp-main.19} {{ETC}:
  Encoding long and structured inputs in transformers}.
\newblock In \emph{Proceedings of the 2020 Conference on Empirical Methods in
  Natural Language Processing (EMNLP)}, pages 268--284, Online. Association for
  Computational Linguistics.

\bibitem[{Beltagy et~al.(2020)Beltagy, Peters, and
  Cohan}]{beltagy2020longformer}
Iz~Beltagy, Matthew~E Peters, and Arman Cohan. 2020.
\newblock Longformer: The long-document transformer.
\newblock \emph{arXiv preprint arXiv:2004.05150}.

\bibitem[{Brown et~al.(2020)Brown, Mann, Ryder, Subbiah, Kaplan, Dhariwal,
  Neelakantan, Shyam, Sastry, Askell, Agarwal, Herbert{-}Voss, Krueger,
  Henighan, Child, Ramesh, Ziegler, Wu, Winter, Hesse, Chen, Sigler, Litwin,
  Gray, Chess, Clark, Berner, McCandlish, Radford, Sutskever, and
  Amodei}]{brown2020language}
Tom~B. Brown, Benjamin Mann, Nick Ryder, Melanie Subbiah, Jared Kaplan,
  Prafulla Dhariwal, Arvind Neelakantan, Pranav Shyam, Girish Sastry, Amanda
  Askell, Sandhini Agarwal, Ariel Herbert{-}Voss, Gretchen Krueger, Tom
  Henighan, Rewon Child, Aditya Ramesh, Daniel~M. Ziegler, Jeffrey Wu, Clemens
  Winter, Christopher Hesse, Mark Chen, Eric Sigler, Mateusz Litwin, Scott
  Gray, Benjamin Chess, Jack Clark, Christopher Berner, Sam McCandlish, Alec
  Radford, Ilya Sutskever, and Dario Amodei. 2020.
\newblock \href
  {https://proceedings.neurips.cc/paper/2020/hash/1457c0d6bfcb4967418bfb8ac142f64a-Abstract.html}
  {Language models are few-shot learners}.
\newblock In \emph{Advances in Neural Information Processing Systems 33: Annual
  Conference on Neural Information Processing Systems 2020, NeurIPS 2020,
  December 6-12, 2020, virtual}.

\bibitem[{Chen et~al.(2016)Chen, Xu, Zhang, and Guestrin}]{chen2016training}
Tianqi Chen, Bing Xu, Chiyuan Zhang, and Carlos Guestrin. 2016.
\newblock Training deep nets with sublinear memory cost.
\newblock \emph{arXiv preprint arXiv:1604.06174}.

\bibitem[{Chen and Bansal(2018)}]{chen-bansal-2018-fast}
Yen-Chun Chen and Mohit Bansal. 2018.
\newblock \href {https://doi.org/10.18653/v1/P18-1063} {Fast abstractive
  summarization with reinforce-selected sentence rewriting}.
\newblock In \emph{Proceedings of the 56th Annual Meeting of the Association
  for Computational Linguistics (Volume 1: Long Papers)}, pages 675--686,
  Melbourne, Australia. Association for Computational Linguistics.

\bibitem[{Cheng and Lapata(2016)}]{cheng-lapata-2016-neural}
Jianpeng Cheng and Mirella Lapata. 2016.
\newblock \href {https://doi.org/10.18653/v1/P16-1046} {Neural summarization by
  extracting sentences and words}.
\newblock In \emph{Proceedings of the 54th Annual Meeting of the Association
  for Computational Linguistics (Volume 1: Long Papers)}, pages 484--494,
  Berlin, Germany. Association for Computational Linguistics.

\bibitem[{Child et~al.(2019)Child, Gray, Radford, and
  Sutskever}]{child2019generating}
Rewon Child, Scott Gray, Alec Radford, and Ilya Sutskever. 2019.
\newblock Generating long sequences with sparse transformers.
\newblock \emph{arXiv preprint arXiv:1904.10509}.

\bibitem[{Cho et~al.(2014)Cho, van Merri{\"e}nboer, Gulcehre, Bahdanau,
  Bougares, Schwenk, and Bengio}]{cho-etal-2014-learning}
Kyunghyun Cho, Bart van Merri{\"e}nboer, Caglar Gulcehre, Dzmitry Bahdanau,
  Fethi Bougares, Holger Schwenk, and Yoshua Bengio. 2014.
\newblock \href {https://doi.org/10.3115/v1/D14-1179} {Learning phrase
  representations using {RNN} encoder{--}decoder for statistical machine
  translation}.
\newblock In \emph{Proceedings of the 2014 Conference on Empirical Methods in
  Natural Language Processing ({EMNLP})}, pages 1724--1734, Doha, Qatar.
  Association for Computational Linguistics.

\bibitem[{Choromanski et~al.(2021)Choromanski, Likhosherstov, Dohan, Song,
  Gane, Sarlos, Hawkins, Davis, Mohiuddin, Kaiser, Belanger, Colwell, and
  Weller}]{choromanski2021rethinking}
Krzysztof~Marcin Choromanski, Valerii Likhosherstov, David Dohan, Xingyou Song,
  Andreea Gane, Tamas Sarlos, Peter Hawkins, Jared~Quincy Davis, Afroz
  Mohiuddin, Lukasz Kaiser, David~Benjamin Belanger, Lucy~J Colwell, and Adrian
  Weller. 2021.
\newblock \href {https://openreview.net/forum?id=Ua6zuk0WRH} {Rethinking
  attention with performers}.
\newblock In \emph{International Conference on Learning Representations}.

\bibitem[{Clifton et~al.(2020)Clifton, Reddy, Yu, Pappu, Rezapour, Bonab,
  Eskevich, Jones, Karlgren, Carterette, and Jones}]{clifton-etal-2020-100000}
Ann Clifton, Sravana Reddy, Yongze Yu, Aasish Pappu, Rezvaneh Rezapour, Hamed
  Bonab, Maria Eskevich, Gareth Jones, Jussi Karlgren, Ben Carterette, and
  Rosie Jones. 2020.
\newblock \href {https://www.aclweb.org/anthology/2020.coling-main.519}
  {100,000 podcasts: A spoken {E}nglish document corpus}.
\newblock In \emph{Proceedings of the 28th International Conference on
  Computational Linguistics}, pages 5903--5917, Barcelona, Spain (Online).
  International Committee on Computational Linguistics.

\bibitem[{Cohan et~al.(2018)Cohan, Dernoncourt, Kim, Bui, Kim, Chang, and
  Goharian}]{cohan-etal-2018-discourse}
Arman Cohan, Franck Dernoncourt, Doo~Soon Kim, Trung Bui, Seokhwan Kim, Walter
  Chang, and Nazli Goharian. 2018.
\newblock \href {https://doi.org/10.18653/v1/N18-2097} {A discourse-aware
  attention model for abstractive summarization of long documents}.
\newblock In \emph{Proceedings of the 2018 Conference of the North {A}merican
  Chapter of the Association for Computational Linguistics: Human Language
  Technologies, Volume 2 (Short Papers)}, pages 615--621, New Orleans,
  Louisiana. Association for Computational Linguistics.

\bibitem[{Dai et~al.(2019)Dai, Yang, Yang, Carbonell, Le, and
  Salakhutdinov}]{dai-etal-2019-transformer}
Zihang Dai, Zhilin Yang, Yiming Yang, Jaime Carbonell, Quoc Le, and Ruslan
  Salakhutdinov. 2019.
\newblock \href {https://doi.org/10.18653/v1/P19-1285} {Transformer-{XL}:
  Attentive language models beyond a fixed-length context}.
\newblock In \emph{Proceedings of the 57th Annual Meeting of the Association
  for Computational Linguistics}, pages 2978--2988, Florence, Italy.
  Association for Computational Linguistics.

\bibitem[{Devlin et~al.(2019)Devlin, Chang, Lee, and
  Toutanova}]{devlin2018bert}
Jacob Devlin, Ming-Wei Chang, Kenton Lee, and Kristina Toutanova. 2019.
\newblock \href {https://doi.org/10.18653/v1/N19-1423} {{BERT}: Pre-training of
  deep bidirectional transformers for language understanding}.
\newblock In \emph{Proceedings of the 2019 Conference of the North {A}merican
  Chapter of the Association for Computational Linguistics: Human Language
  Technologies, Volume 1 (Long and Short Papers)}, pages 4171--4186,
  Minneapolis, Minnesota. Association for Computational Linguistics.

\bibitem[{Dou et~al.(2021)Dou, Liu, Hayashi, Jiang, and
  Neubig}]{dou-etal-2021-gsum}
Zi-Yi Dou, Pengfei Liu, Hiroaki Hayashi, Zhengbao Jiang, and Graham Neubig.
  2021.
\newblock \href {https://www.aclweb.org/anthology/2021.naacl-main.384} {{GS}um:
  A general framework for guided neural abstractive summarization}.
\newblock In \emph{Proceedings of the 2021 Conference of the North American
  Chapter of the Association for Computational Linguistics: Human Language
  Technologies}, pages 4830--4842, Online. Association for Computational
  Linguistics.

\bibitem[{Gehrmann et~al.(2018)Gehrmann, Deng, and
  Rush}]{gehrmann-etal-2018-bottom}
Sebastian Gehrmann, Yuntian Deng, and Alexander Rush. 2018.
\newblock \href {https://doi.org/10.18653/v1/D18-1443} {Bottom-up abstractive
  summarization}.
\newblock In \emph{Proceedings of the 2018 Conference on Empirical Methods in
  Natural Language Processing}, pages 4098--4109, Brussels, Belgium.
  Association for Computational Linguistics.

\bibitem[{Gidiotis and Tsoumakas(2020)}]{gidiotis2020divide}
Alexios Gidiotis and Grigorios Tsoumakas. 2020.
\newblock \href {https://doi.org/10.1109/TASLP.2020.3037401} {A
  divide-and-conquer approach to the summarization of long documents}.
\newblock \emph{IEEE/ACM Transactions on Audio, Speech, and Language
  Processing}, 28:3029--3040.

\bibitem[{He et~al.(2020)He, Kry{\'s}ci{\'n}ski, McCann, Rajani, and
  Xiong}]{he2020ctrlsum}
Junxian He, Wojciech Kry{\'s}ci{\'n}ski, Bryan McCann, Nazneen Rajani, and
  Caiming Xiong. 2020.
\newblock {CTRL}sum: Towards generic controllable text summarization.
\newblock \emph{arXiv preprint arXiv:2012.04281}.

\bibitem[{Hermann et~al.(2015)Hermann, Kocisk{\'{y}}, Grefenstette, Espeholt,
  Kay, Suleyman, and Blunsom}]{hermann2015teaching}
Karl~Moritz Hermann, Tom{\'{a}}s Kocisk{\'{y}}, Edward Grefenstette, Lasse
  Espeholt, Will Kay, Mustafa Suleyman, and Phil Blunsom. 2015.
\newblock \href
  {https://proceedings.neurips.cc/paper/2015/hash/afdec7005cc9f14302cd0474fd0f3c96-Abstract.html}
  {Teaching machines to read and comprehend}.
\newblock In \emph{Advances in Neural Information Processing Systems 28: Annual
  Conference on Neural Information Processing Systems 2015, December 7-12,
  2015, Montreal, Quebec, Canada}, pages 1693--1701.

\bibitem[{Hinton et~al.(2015)Hinton, Vinyals, and Dean}]{hinton2015distilling}
Geoffrey Hinton, Oriol Vinyals, and Jeffrey Dean. 2015.
\newblock \href {http://arxiv.org/abs/1503.02531} {Distilling the knowledge in
  a neural network}.
\newblock In \emph{NIPS Deep Learning and Representation Learning Workshop}.

\bibitem[{Hsu et~al.(2018)Hsu, Lin, Lee, Min, Tang, and
  Sun}]{hsu-etal-2018-unified}
Wan-Ting Hsu, Chieh-Kai Lin, Ming-Ying Lee, Kerui Min, Jing Tang, and Min Sun.
  2018.
\newblock \href {https://doi.org/10.18653/v1/P18-1013} {A unified model for
  extractive and abstractive summarization using inconsistency loss}.
\newblock In \emph{Proceedings of the 56th Annual Meeting of the Association
  for Computational Linguistics (Volume 1: Long Papers)}, pages 132--141,
  Melbourne, Australia. Association for Computational Linguistics.

\bibitem[{Huang et~al.(2019)Huang, Cheng, Bapna, Firat, Chen, Chen, Lee, Ngiam,
  Le, Wu, and Chen}]{huang2019gpipe}
Yanping Huang, Youlong Cheng, Ankur Bapna, Orhan Firat, Dehao Chen, Mia~Xu
  Chen, HyoukJoong Lee, Jiquan Ngiam, Quoc~V. Le, Yonghui Wu, and Zhifeng Chen.
  2019.
\newblock \href
  {https://proceedings.neurips.cc/paper/2019/hash/093f65e080a295f8076b1c5722a46aa2-Abstract.html}
  {Gpipe: Efficient training of giant neural networks using pipeline
  parallelism}.
\newblock In \emph{Advances in Neural Information Processing Systems 32: Annual
  Conference on Neural Information Processing Systems 2019, NeurIPS 2019,
  December 8-14, 2019, Vancouver, BC, Canada}, pages 103--112.

\bibitem[{Jones et~al.(2020)Jones, Carterette, Clifton, Eskevich, Jones,
  Karlgren, Pappu, Reddy, and Yu}]{jones_trec2020}
Rosie Jones, Ben Carterette, Ann Clifton, Maria Eskevich, Gareth J.~F. Jones,
  Jussi Karlgren, Aasish Pappu, Sravana Reddy, and Yongze Yu. 2020.
\newblock Trec 2020 podcasts track overview.
\newblock In \emph{The 29th Text Retrieval Conference (TREC) notebook}.

\bibitem[{Kingma and Ba(2015)}]{kingma2014adam}
Diederik~P. Kingma and Jimmy Ba. 2015.
\newblock \href {http://arxiv.org/abs/1412.6980} {Adam: {A} method for
  stochastic optimization}.
\newblock In \emph{3rd International Conference on Learning Representations,
  {ICLR} 2015, San Diego, CA, USA, May 7-9, 2015, Conference Track
  Proceedings}.

\bibitem[{Kitaev et~al.(2020)Kitaev, Kaiser, and Levskaya}]{kitaev2020reformer}
Nikita Kitaev, Lukasz Kaiser, and Anselm Levskaya. 2020.
\newblock \href {https://openreview.net/forum?id=rkgNKkHtvB} {Reformer: The
  efficient transformer}.
\newblock In \emph{8th International Conference on Learning Representations,
  {ICLR} 2020, Addis Ababa, Ethiopia, April 26-30, 2020}. OpenReview.net.

\bibitem[{Lebanoff et~al.(2020)Lebanoff, Dernoncourt, Kim, Chang, and
  Liu}]{lebanoff-etal-2020-cascade}
Logan Lebanoff, Franck Dernoncourt, Doo~Soon Kim, Walter Chang, and Fei Liu.
  2020.
\newblock \href {https://www.aclweb.org/anthology/2020.aacl-main.52} {A cascade
  approach to neural abstractive summarization with content selection and
  fusion}.
\newblock In \emph{Proceedings of the 1st Conference of the Asia-Pacific
  Chapter of the Association for Computational Linguistics and the 10th
  International Joint Conference on Natural Language Processing}, pages
  529--535, Suzhou, China. Association for Computational Linguistics.

\bibitem[{Lewis et~al.(2020)Lewis, Liu, Goyal, Ghazvininejad, Mohamed, Levy,
  Stoyanov, and Zettlemoyer}]{lewis-etal-2020-bart}
Mike Lewis, Yinhan Liu, Naman Goyal, Marjan Ghazvininejad, Abdelrahman Mohamed,
  Omer Levy, Veselin Stoyanov, and Luke Zettlemoyer. 2020.
\newblock \href {https://doi.org/10.18653/v1/2020.acl-main.703} {{BART}:
  Denoising sequence-to-sequence pre-training for natural language generation,
  translation, and comprehension}.
\newblock In \emph{Proceedings of the 58th Annual Meeting of the Association
  for Computational Linguistics}, pages 7871--7880, Online. Association for
  Computational Linguistics.

\bibitem[{Li et~al.(2019)Li, Zhang, Ji, and Radke}]{li-etal-2019-keep}
Manling Li, Lingyu Zhang, Heng Ji, and Richard~J. Radke. 2019.
\newblock \href {https://doi.org/10.18653/v1/P19-1210} {Keep meeting summaries
  on topic: Abstractive multi-modal meeting summarization}.
\newblock In \emph{Proceedings of the 57th Annual Meeting of the Association
  for Computational Linguistics}, pages 2190--2196, Florence, Italy.
  Association for Computational Linguistics.

\bibitem[{Lin(2004)}]{lin-2004-rouge}
Chin-Yew Lin. 2004.
\newblock \href {https://www.aclweb.org/anthology/W04-1013} {{ROUGE}: A package
  for automatic evaluation of summaries}.
\newblock In \emph{Text Summarization Branches Out}, pages 74--81, Barcelona,
  Spain. Association for Computational Linguistics.

\bibitem[{Liu et~al.(2018)Liu, Saleh, Pot, Goodrich, Sepassi, Kaiser, and
  Shazeer}]{liu2018generating}
Peter~J. Liu, Mohammad Saleh, Etienne Pot, Ben Goodrich, Ryan Sepassi, Lukasz
  Kaiser, and Noam Shazeer. 2018.
\newblock \href {https://openreview.net/forum?id=Hyg0vbWC-} {Generating
  wikipedia by summarizing long sequences}.
\newblock In \emph{6th International Conference on Learning Representations,
  {ICLR} 2018, Vancouver, BC, Canada, April 30 - May 3, 2018, Conference Track
  Proceedings}. OpenReview.net.

\bibitem[{Liu and Lapata(2019{\natexlab{a}})}]{liu-lapata-2019-hierarchical}
Yang Liu and Mirella Lapata. 2019{\natexlab{a}}.
\newblock \href {https://doi.org/10.18653/v1/P19-1500} {Hierarchical
  transformers for multi-document summarization}.
\newblock In \emph{Proceedings of the 57th Annual Meeting of the Association
  for Computational Linguistics}, pages 5070--5081, Florence, Italy.
  Association for Computational Linguistics.

\bibitem[{Liu and Lapata(2019{\natexlab{b}})}]{liu-lapata-2019-text}
Yang Liu and Mirella Lapata. 2019{\natexlab{b}}.
\newblock \href {https://doi.org/10.18653/v1/D19-1387} {Text summarization with
  pretrained encoders}.
\newblock In \emph{Proceedings of the 2019 Conference on Empirical Methods in
  Natural Language Processing and the 9th International Joint Conference on
  Natural Language Processing (EMNLP-IJCNLP)}, pages 3730--3740, Hong Kong,
  China. Association for Computational Linguistics.

\bibitem[{Manakul and Gales(2020)}]{manakul2020cued_speech}
Potsawee Manakul and Mark Gales. 2020.
\newblock {CUED}\_speech at {TREC} 2020 podcast summarisation track.
\newblock \emph{arXiv preprint arXiv:2012.02535}.

\bibitem[{Manakul et~al.(2020)Manakul, Gales, and
  Wang}]{manakul2020_interspeech}
Potsawee Manakul, Mark~J.F. Gales, and Linlin Wang. 2020.
\newblock \href {https://doi.org/10.21437/Interspeech.2020-1683} {{Abstractive
  Spoken Document Summarization Using Hierarchical Model with Multi-Stage
  Attention Diversity Optimization}}.
\newblock In \emph{Proc. Interspeech 2020}, pages 4248--4252.

\bibitem[{Michel et~al.(2019)Michel, Levy, and Neubig}]{michel_sixteen_heads}
Paul Michel, Omer Levy, and Graham Neubig. 2019.
\newblock \href
  {https://proceedings.neurips.cc/paper/2019/hash/2c601ad9d2ff9bc8b282670cdd54f69f-Abstract.html}
  {Are sixteen heads really better than one?}
\newblock In \emph{Advances in Neural Information Processing Systems 32: Annual
  Conference on Neural Information Processing Systems 2019, NeurIPS 2019,
  December 8-14, 2019, Vancouver, BC, Canada}, pages 14014--14024.

\bibitem[{Micikevicius et~al.(2018)Micikevicius, Narang, Alben, Diamos, Elsen,
  Garc{\'{\i}}a, Ginsburg, Houston, Kuchaiev, Venkatesh, and
  Wu}]{micikevicius2017mixed}
Paulius Micikevicius, Sharan Narang, Jonah Alben, Gregory~F. Diamos, Erich
  Elsen, David Garc{\'{\i}}a, Boris Ginsburg, Michael Houston, Oleksii
  Kuchaiev, Ganesh Venkatesh, and Hao Wu. 2018.
\newblock \href {https://openreview.net/forum?id=r1gs9JgRZ} {Mixed precision
  training}.
\newblock In \emph{6th International Conference on Learning Representations,
  {ICLR} 2018, Vancouver, BC, Canada, April 30 - May 3, 2018, Conference Track
  Proceedings}. OpenReview.net.

\bibitem[{Nallapati et~al.(2016)Nallapati, Zhou, dos Santos, Gu̇l{\c{c}}ehre,
  and Xiang}]{nallapati-etal-2016-abstractive}
Ramesh Nallapati, Bowen Zhou, Cicero dos Santos, {\c{C}}a{\u{g}}lar
  Gu̇l{\c{c}}ehre, and Bing Xiang. 2016.
\newblock \href {https://doi.org/10.18653/v1/K16-1028} {Abstractive text
  summarization using sequence-to-sequence {RNN}s and beyond}.
\newblock In \emph{Proceedings of The 20th {SIGNLL} Conference on Computational
  Natural Language Learning}, pages 280--290, Berlin, Germany. Association for
  Computational Linguistics.

\bibitem[{Narayan et~al.(2020)Narayan, Maynez, Adamek, Pighin, Bratanic, and
  McDonald}]{narayan-etal-2020-stepwise}
Shashi Narayan, Joshua Maynez, Jakub Adamek, Daniele Pighin, Blaz Bratanic, and
  Ryan McDonald. 2020.
\newblock \href {https://doi.org/10.18653/v1/2020.emnlp-main.339} {Stepwise
  extractive summarization and planning with structured transformers}.
\newblock In \emph{Proceedings of the 2020 Conference on Empirical Methods in
  Natural Language Processing (EMNLP)}, pages 4143--4159, Online. Association
  for Computational Linguistics.

\bibitem[{Parmar et~al.(2018)Parmar, Vaswani, Uszkoreit, Kaiser, Shazeer, Ku,
  and Tran}]{parmar2018image}
Niki Parmar, Ashish Vaswani, Jakob Uszkoreit, Lukasz Kaiser, Noam Shazeer,
  Alexander Ku, and Dustin Tran. 2018.
\newblock \href {http://proceedings.mlr.press/v80/parmar18a.html} {Image
  transformer}.
\newblock In \emph{Proceedings of the 35th International Conference on Machine
  Learning, {ICML} 2018, Stockholmsm{\"{a}}ssan, Stockholm, Sweden, July 10-15,
  2018}, volume~80 of \emph{Proceedings of Machine Learning Research}, pages
  4052--4061. {PMLR}.

\bibitem[{Pilault et~al.(2020)Pilault, Li, Subramanian, and
  Pal}]{pilault-etal-2020-extractive}
Jonathan Pilault, Raymond Li, Sandeep Subramanian, and Chris Pal. 2020.
\newblock \href {https://doi.org/10.18653/v1/2020.emnlp-main.748} {On
  extractive and abstractive neural document summarization with transformer
  language models}.
\newblock In \emph{Proceedings of the 2020 Conference on Empirical Methods in
  Natural Language Processing (EMNLP)}, pages 9308--9319, Online. Association
  for Computational Linguistics.

\bibitem[{Qiu et~al.(2020)Qiu, Ma, Levy, Yih, Wang, and
  Tang}]{qiu-etal-2020-blockwise}
Jiezhong Qiu, Hao Ma, Omer Levy, Wen-tau Yih, Sinong Wang, and Jie Tang. 2020.
\newblock \href {https://doi.org/10.18653/v1/2020.findings-emnlp.232}
  {Blockwise self-attention for long document understanding}.
\newblock In \emph{Findings of the Association for Computational Linguistics:
  EMNLP 2020}, pages 2555--2565, Online. Association for Computational
  Linguistics.

\bibitem[{Radford et~al.(2019)Radford, Wu, Child, Luan, Amodei, and
  Sutskever}]{radford2019language}
Alec Radford, Jeffrey Wu, Rewon Child, David Luan, Dario Amodei, and Ilya
  Sutskever. 2019.
\newblock Language models are unsupervised multitask learners.
\newblock \emph{OpenAI blog}, 1(8):9.

\bibitem[{Raffel et~al.(2020)Raffel, Shazeer, Roberts, Lee, Narang, Matena,
  Zhou, Li, and Liu}]{raffel2020exploring}
Colin Raffel, Noam Shazeer, Adam Roberts, Katherine Lee, Sharan Narang, Michael
  Matena, Yanqi Zhou, Wei Li, and Peter~J Liu. 2020.
\newblock Exploring the limits of transfer learning with a unified text-to-text
  transformer.
\newblock \emph{Journal of Machine Learning Research}.

\bibitem[{Sanh et~al.(2019)Sanh, Debut, Chaumond, and
  Wolf}]{sanh2019distilbert}
Victor Sanh, Lysandre Debut, Julien Chaumond, and Thomas Wolf. 2019.
\newblock Distilbert, a distilled version of bert: smaller, faster, cheaper and
  lighter.
\newblock \emph{arXiv preprint arXiv:1910.01108}.

\bibitem[{See et~al.(2017)See, Liu, and Manning}]{see-etal-2017-get}
Abigail See, Peter~J. Liu, and Christopher~D. Manning. 2017.
\newblock \href {https://doi.org/10.18653/v1/P17-1099} {Get to the point:
  Summarization with pointer-generator networks}.
\newblock In \emph{Proceedings of the 55th Annual Meeting of the Association
  for Computational Linguistics (Volume 1: Long Papers)}, pages 1073--1083,
  Vancouver, Canada. Association for Computational Linguistics.

\bibitem[{Song et~al.(2020)Song, Li, Wang, Yu, and Liu}]{kaiqiang_trec2020}
Kaiqiang Song, Chen Li, Xiaoyang Wang, Dong Yu, and Fei Liu. 2020.
\newblock Automatic summarization of open-domain podcast episodes.
\newblock \emph{arXiv preprint arXiv:2011.04132}.

\bibitem[{Tay et~al.(2020{\natexlab{a}})Tay, Bahri, Yang, Metzler, and
  Juan}]{tay2020sparse}
Yi~Tay, Dara Bahri, Liu Yang, Donald Metzler, and Da{-}Cheng Juan.
  2020{\natexlab{a}}.
\newblock \href {http://proceedings.mlr.press/v119/tay20a.html} {Sparse
  sinkhorn attention}.
\newblock In \emph{Proceedings of the 37th International Conference on Machine
  Learning, {ICML} 2020, 13-18 July 2020, Virtual Event}, volume 119 of
  \emph{Proceedings of Machine Learning Research}, pages 9438--9447. {PMLR}.

\bibitem[{Tay et~al.(2021)Tay, Dehghani, Abnar, Shen, Bahri, Pham, Rao, Yang,
  Ruder, and Metzler}]{tay2021long}
Yi~Tay, Mostafa Dehghani, Samira Abnar, Yikang Shen, Dara Bahri, Philip Pham,
  Jinfeng Rao, Liu Yang, Sebastian Ruder, and Donald Metzler. 2021.
\newblock \href {https://openreview.net/forum?id=qVyeW-grC2k} {Long range arena
  : A benchmark for efficient transformers}.
\newblock In \emph{International Conference on Learning Representations}.

\bibitem[{Tay et~al.(2020{\natexlab{b}})Tay, Dehghani, Bahri, and
  Metzler}]{tay2020efficient}
Yi~Tay, Mostafa Dehghani, Dara Bahri, and Donald Metzler. 2020{\natexlab{b}}.
\newblock Efficient transformers: A survey.
\newblock \emph{arXiv preprint arXiv:2009.06732}.

\bibitem[{Vaswani et~al.(2017)Vaswani, Shazeer, Parmar, Uszkoreit, Jones,
  Gomez, Kaiser, and Polosukhin}]{vaswani2017attention}
Ashish Vaswani, Noam Shazeer, Niki Parmar, Jakob Uszkoreit, Llion Jones,
  Aidan~N. Gomez, Lukasz Kaiser, and Illia Polosukhin. 2017.
\newblock \href
  {https://proceedings.neurips.cc/paper/2017/hash/3f5ee243547dee91fbd053c1c4a845aa-Abstract.html}
  {Attention is all you need}.
\newblock In \emph{Advances in Neural Information Processing Systems 30: Annual
  Conference on Neural Information Processing Systems 2017, December 4-9, 2017,
  Long Beach, CA, {USA}}, pages 5998--6008.

\bibitem[{Voita et~al.(2019)Voita, Talbot, Moiseev, Sennrich, and
  Titov}]{voita-etal-2019-analyzing}
Elena Voita, David Talbot, Fedor Moiseev, Rico Sennrich, and Ivan Titov. 2019.
\newblock \href {https://doi.org/10.18653/v1/P19-1580} {Analyzing multi-head
  self-attention: Specialized heads do the heavy lifting, the rest can be
  pruned}.
\newblock In \emph{Proceedings of the 57th Annual Meeting of the Association
  for Computational Linguistics}, pages 5797--5808, Florence, Italy.
  Association for Computational Linguistics.

\bibitem[{Wang et~al.(2020)Wang, Li, Khabsa, Fang, and Ma}]{wang2020linformer}
Sinong Wang, Belinda Li, Madian Khabsa, Han Fang, and Hao Ma. 2020.
\newblock Linformer: Self-attention with linear complexity.
\newblock \emph{arXiv preprint arXiv:2006.04768}.

\bibitem[{Wolf et~al.(2020)Wolf, Debut, Sanh, Chaumond, Delangue, Moi, Cistac,
  Rault, Louf, Funtowicz, Davison, Shleifer, von Platen, Ma, Jernite, Plu, Xu,
  Le~Scao, Gugger, Drame, Lhoest, and Rush}]{wolf-etal-2020-transformers}
Thomas Wolf, Lysandre Debut, Victor Sanh, Julien Chaumond, Clement Delangue,
  Anthony Moi, Pierric Cistac, Tim Rault, Remi Louf, Morgan Funtowicz, Joe
  Davison, Sam Shleifer, Patrick von Platen, Clara Ma, Yacine Jernite, Julien
  Plu, Canwen Xu, Teven Le~Scao, Sylvain Gugger, Mariama Drame, Quentin Lhoest,
  and Alexander Rush. 2020.
\newblock \href {https://doi.org/10.18653/v1/2020.emnlp-demos.6} {Transformers:
  State-of-the-art natural language processing}.
\newblock In \emph{Proceedings of the 2020 Conference on Empirical Methods in
  Natural Language Processing: System Demonstrations}, pages 38--45, Online.
  Association for Computational Linguistics.

\bibitem[{Xiao and Carenini(2019)}]{xiao-carenini-2019-extractive}
Wen Xiao and Giuseppe Carenini. 2019.
\newblock \href {https://doi.org/10.18653/v1/D19-1298} {Extractive
  summarization of long documents by combining global and local context}.
\newblock In \emph{Proceedings of the 2019 Conference on Empirical Methods in
  Natural Language Processing and the 9th International Joint Conference on
  Natural Language Processing (EMNLP-IJCNLP)}, pages 3011--3021, Hong Kong,
  China. Association for Computational Linguistics.

\bibitem[{Xiao and Carenini(2020)}]{xiao-carenini-2020-systematically}
Wen Xiao and Giuseppe Carenini. 2020.
\newblock \href {https://www.aclweb.org/anthology/2020.aacl-main.51}
  {Systematically exploring redundancy reduction in summarizing long
  documents}.
\newblock In \emph{Proceedings of the 1st Conference of the Asia-Pacific
  Chapter of the Association for Computational Linguistics and the 10th
  International Joint Conference on Natural Language Processing}, pages
  516--528, Suzhou, China. Association for Computational Linguistics.

\bibitem[{Zaheer et~al.(2020)Zaheer, Guruganesh, Dubey, Ainslie, Alberti,
  Ontanon, Pham, Ravula, Wang, Yang et~al.}]{zaheer2020big}
Manzil Zaheer, Guru Guruganesh, Kumar~Avinava Dubey, Joshua Ainslie, Chris
  Alberti, Santiago Ontanon, Philip Pham, Anirudh Ravula, Qifan Wang, Li~Yang,
  et~al. 2020.
\newblock Big bird: Transformers for longer sequences.
\newblock \emph{Advances in Neural Information Processing Systems}, 33.

\bibitem[{Zhang et~al.(2020)Zhang, Zhao, Saleh, and Liu}]{zhang2020pegasus}
Jingqing Zhang, Yao Zhao, Mohammad Saleh, and Peter Liu. 2020.
\newblock Pegasus: Pre-training with extracted gap-sentences for abstractive
  summarization.
\newblock In \emph{International Conference on Machine Learning}, pages
  11328--11339. PMLR.

\bibitem[{Zhang et~al.(2019)Zhang, Wei, and Zhou}]{zhang-etal-2019-hibert}
Xingxing Zhang, Furu Wei, and Ming Zhou. 2019.
\newblock \href {https://doi.org/10.18653/v1/P19-1499} {{HIBERT}: Document
  level pre-training of hierarchical bidirectional transformers for document
  summarization}.
\newblock In \emph{Proceedings of the 57th Annual Meeting of the Association
  for Computational Linguistics}, pages 5059--5069, Florence, Italy.
  Association for Computational Linguistics.

\bibitem[{Zhao et~al.(2019)Zhao, Pan, Fan, Liu, Li, and Yang}]{hier_rnn_2019}
Zhou Zhao, Haojie Pan, Changjie Fan, Yan Liu, Linlin Li, and Min Yang. 2019.
\newblock \href {https://doi.org/10.1145/3308558.3313619} {Abstractive meeting
  summarization via hierarchical adaptive segmental network learning}.
\newblock In \emph{The World Wide Web Conference, {WWW} 2019, San Francisco,
  CA, USA, May 13-17, 2019}, pages 3455--3461. {ACM}.

\end{thebibliography}

\appendix

\newpage
\section{Detailed Memory \& Time Analysis}
\label{appendix:complexity}
Our memory analysis is system-independent, albeit implementation-dependent. We carry out the experiments using PyTorch version 1.2.0. We use \verb|pytorch_memlab|\footnote{\url{https://github.com/Stonesjtu/pytorch_memlab}} to compute GPU memory during forward and backward passes. Our notation is: input length $N$, target length $M$, local self-attention width $W$, and batch size $B$.

\subsection{BART Memory}
 We collect 30 samples, spanning $N \in [64,3000]$ and $M \in [36,576]$ using batch size of 1. Our least-squared regression of the memory equation $\texttt{memory} = c^b_1 +  B(c^b_2 M + c^b_3 N + c^b_4 MN + c^b_5 M^2 + c^b_6 N^2)$ yields $R^2=1, \text{RMSE}=0.026$, and the coefficients are:  $c^b_1=6.054$, $c^b_2=1.594\times10^{-3}$, $c^b_3=8.192\times10^{-4}$, $c^b_4=1.418\times10^{-6}$, $c^b_5=1.077\times10^{-6}$, $c^b_6=1.456\times10^{-6}$.
\subsubsection*{Model and Optimizer}
The \textit{constant} term $c^b_1=6.054$ GiB is independent of batch size, system, or implementation (given the same floating-point precision). This term comprises model and optimizer memory as follows (in 32-bit floating point, 1 variable takes 4 bytes):
\begin{enumerate}
    \item Model Parameter: BART has 406,290,432 parameters, yielding $406290432 \times 4 = 1.625\times10^9  \text{bytes} =$ 1.51 GiB.
    \item Model Gradient: Each parameter has one corresponding gradient variable, e.g. \verb|.grad| in PyTorch. Thus, this also occupies 1.51 GiB.
    \item Optimizer: Adam optimizer \cite{kingma2014adam} stores first moment and second moment for each and every model parameters, hence, taking 3.02 GiB.
\end{enumerate}

\subsubsection*{Activation}
The terms corresponding to $c^b_2,...,c^b_6$ are associated with activation buffers cached for computing gradients in backpropagation. These terms grow linearly with batch size. The dominant term $c^b_6N^2B$ grows quadratically with the input length $N$, motivating encoder's local self-attention design.


\citet{chen2016training} proposes a method to save the activation memory by only caching buffers of a subset of layers, and re-computing the rest dynamically during backpropagation. This results in repeated computations and more training time.

\subsection{LoBART Memory}
We collect 36 samples, spanning $N \in [512, 4096]$, $M \in [100, 400]$, and $W \in [32, 512]$ using batch size of 1. Our least-squared regression of the memory equation $\texttt{memory} = c^l_1 +  B(c^l_2 M + c^l_3 N + c^l_4 MN + c^l_5 M^2 + c^l_6 NW)$ yields $\text{RMSE}=0.010$, and the coefficients are: $c^l_1=6.104$, $c^l_2=1.443\times10^{-3}$, $c^l_3=1.032\times10^{-3}$, $c^l_4=1.487\times10^{-6}$, $c^l_5=1.277\times10^{-6}$, $c^l_6=2.503\times10^{-6}$. The model and optimizer memory is similar to the analysis for BART. The activation memory is now dominated by $c^l_6NW \times B$, where $c^l_6 = 1.72c^b_6$. Thus, we highlight that once $W>0.58N$, LoBART no longer reduces memory. Note that we also tried incorporating the terms $N^2$ and $W$ in the least-squared regression analysis, but their resulting coefficients are small, making both terms negligible. This is expected as quadratic self-attention is replaced by local attention of width $W$, and the width $W$ only determines the receptive field of each and every position in $N$, resulting in the $NW$ term. 
\subsection{Time: BART \& LoBART}
\begin{figure}[ht]
    \centering
      \includegraphics[width=0.80\linewidth,keepaspectratio]{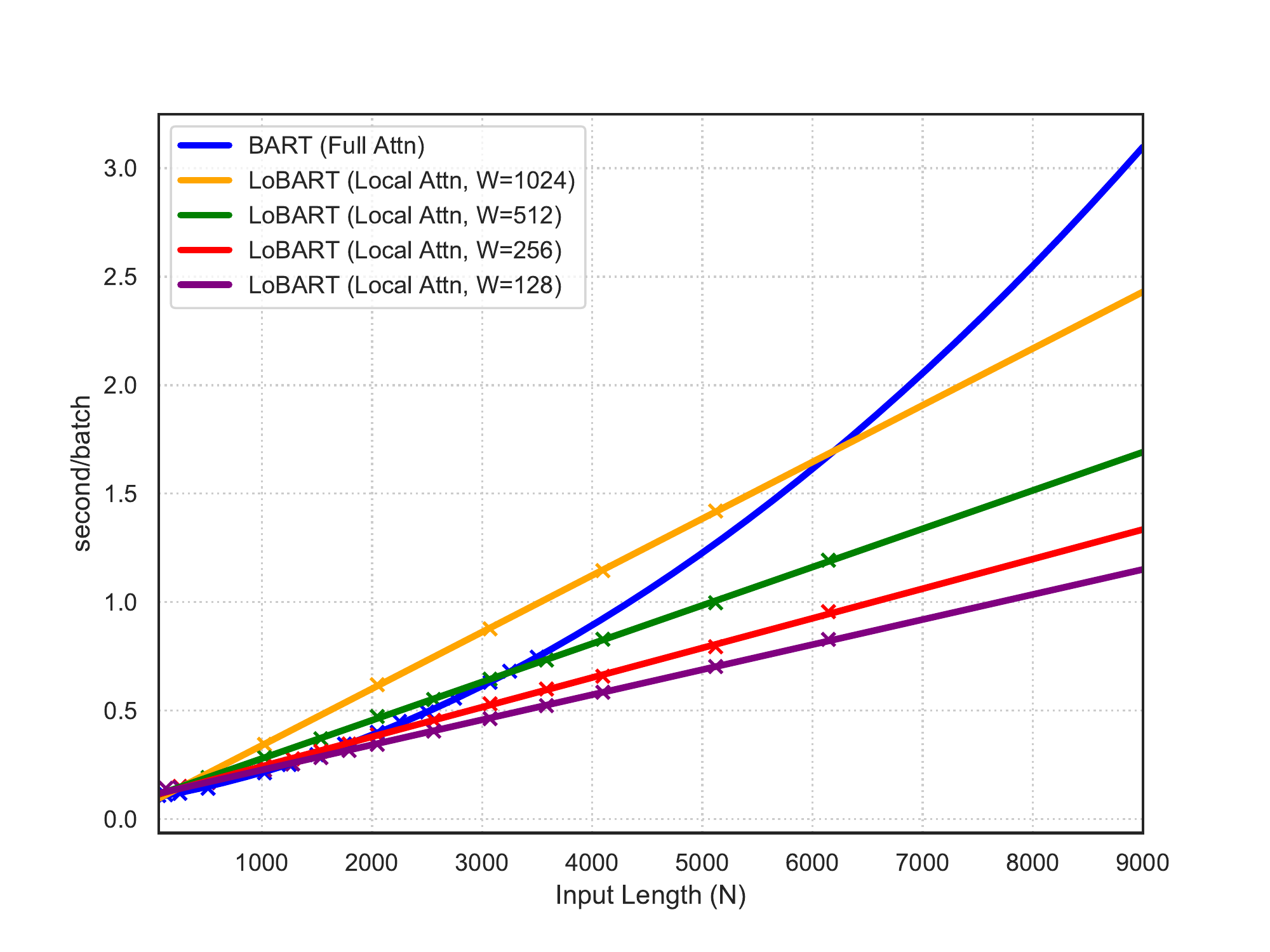}
    \caption{Quadratic time complexity of BART and Linear time complexity of LoBART for $M$=144 and $B$=1.}
    \label{fig:time_pred}
\end{figure}

Unlike memory, time requirement is both system and implementation dependent. In this analysis, we show the results on our infrastructure consisting of a 32 GiB V100 GPU and 32-core Intel Xeon 4215R CPU (3.20GHz). We compute the time required for 50 forward and backward passes in 12 settings for each model configuration. Similar to the memory analysis, we perform least-squared regression where the results are shown in Figure \ref{fig:time_pred}. It can be seen that although LoBART reduces memory requirement, when it comes to time requirement, LoBART is only comparable to BART. This is due to the implementation of local self-attention that involves more processes such as chunking.

\newpage
\section{Implementation Details}
\label{appendix:implementation_details}
\subsection{Models}
\label{appendix:model_parameters}
\textbf{BART \& LoBART.} \\
\noindent We use publicly released BART-large.\footnote{\url{https://huggingface.co/facebook/bart-large-cnn}} For LoBART, our local self-attention is based on HuggingFace's implementation \cite{wolf-etal-2020-transformers}.\footnote{\url{https://huggingface.co/transformers/}} The number of parameters in BART is 406M.

\begin{figure}[ht]
    \centering
      \includegraphics[width=0.70\linewidth,keepaspectratio]{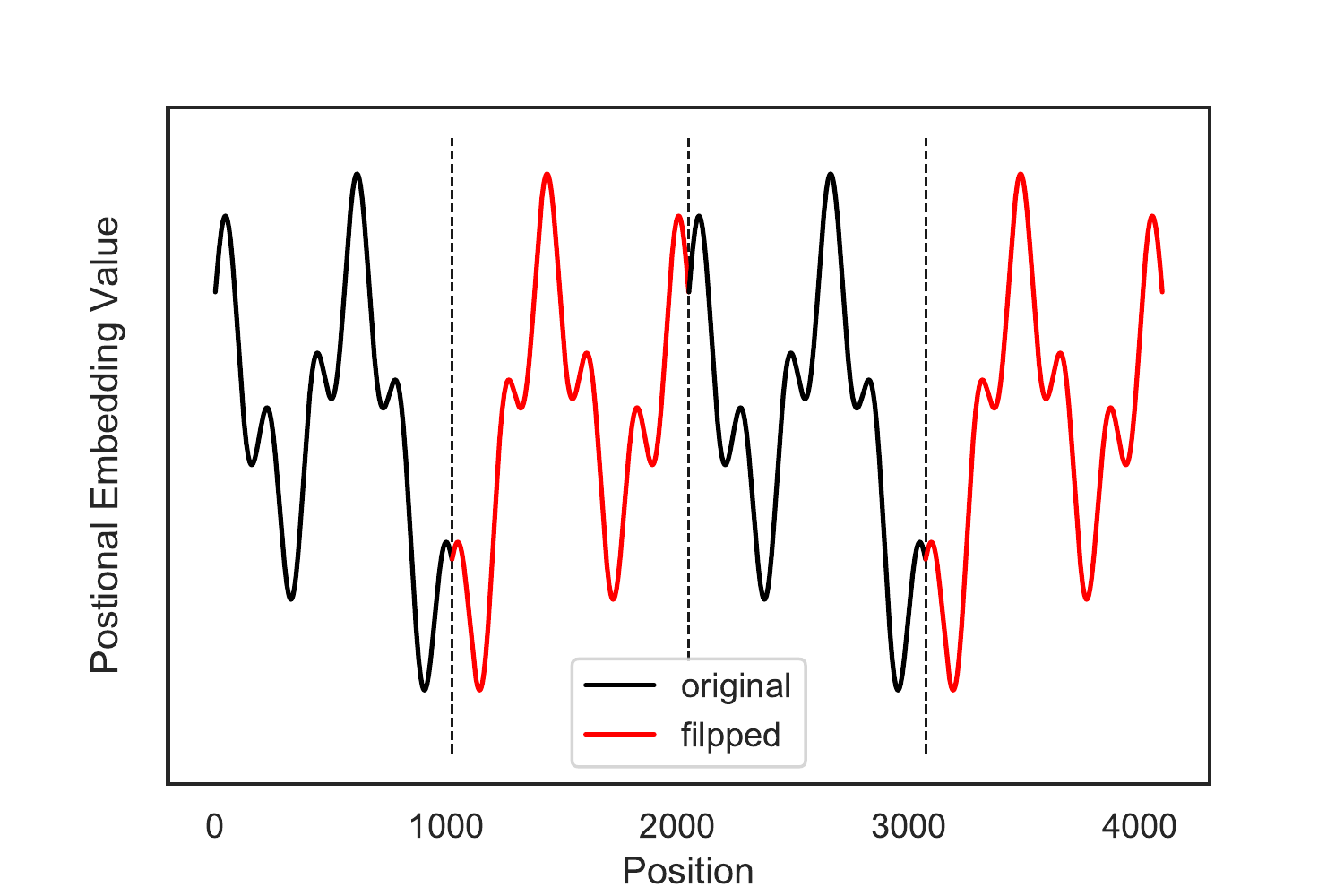}
    \caption{LoBART positional embedding is initialized by copying and flipping BART's positional embedding.}
    \label{fig:positional_embedding}
\end{figure}

The positional embedding of LoBART beyond 1,024 is created by copying BART's positional embedding with flipping to allow a smoother transition as shown in Figure \ref{fig:positional_embedding}, and the number of parameters in LoBART($n$k) is 406M + 50,264$\times$($n$-1)$\times$1,024.

\noindent \textbf{Hierarchical RNN.} \\
\noindent The encoder consists of word-level and sentence-level bidirectional GRUs. The word-level GRU takes embedding vector $\mathbf{e}_{i,j}$ of word $i$ in sentence $j$, and outputs forward representation $\mathbf{h}^{\tt(f)}_{i,j}$ and backward representation $\mathbf{h}^{\tt(b)}_{i,j}$. The sentence-level GRU takes concatenated vector [$\mathbf{h}^{\tt(f)}_{N_j,j}$;$\mathbf{h}^{\tt(b)}_{1,j}$], and outputs sentence representation $\mathbf{h}_j$. The decoder consists of a unidirectional GRU. Each of the encoder GRUs has 2 layers with a dropout layer (p=0.1), and the decoder GRU has 1 layer.  There are word-level and sentence-level attention mechanisms connecting the encoder and decoder. The classification head is a single-layer feedforward layer. The dimension of embedding space is 256, and the hidden size is 512. The number of parameters is 52M.

\subsection{Training \& Inference Hyperparameters}
We process data using the same byte-pair-encoding tokenizer as the BART-large tokenizer, and we use NLTK tokenizer for sentence splitting. We use 32-bit precision training. We stop training when the loss on the validation set stop improving for 3 times. For example, the training steps are approximately: 180k for Podcast; 240k for arXiv; 160k for PubMed. We report the validation performance when training is stopped in Table \ref{tab:valid_performance}. Adam optimizer is used for all experiments with learning rate:
\begin{equation*}
    \text{lr} = 0.002 \times \text{min}(\text{step}^{-0.5}, \text{step}\times\text{warmup}^{-1.5})
\end{equation*}

\begin{table}[ht]
  \centering
  \scalebox{0.9}{
  \begin{tabular}{lrr}
    \toprule
    Parameter &Podcast &arXiv/PubMed\\
    \midrule
    max. tgt len $M$ &144    &400\\
    dropout   &0.1  &0.1 \\
    batch size &1  &1 \\
    gradient accum. &2 &2 \\
    warmup     &10,000 &20,000 \\
    valid step &20,000 &20,000 \\
    \midrule
    loss      &\multicolumn{2}{r}{cross entropy} \\
    compute (BART)       &  \multicolumn{2}{r}{{1$\times$GTX TITAN X (12GiB)}} \\
    compute (LoBART)     &  \multicolumn{2}{r}{{1$\times$V100 (32GiB)}} \\
    \bottomrule
  \end{tabular}}
  \caption{BART/LoBART Training Hyperparameters.}
\end{table}

\begin{table}[ht]
\tabcolsep=0.07cm
  \centering
   \scalebox{0.9}{
  \begin{tabular}{lrr}
    \toprule
    Parameter &Podcast &arXiv/PubMed\\
    \midrule
    max. src \#sent                &1000    &640\\
    max. src \#words-in-sent     &50     &120\\
    max. tgt len $M$           &144    &400\\
    dropout   &0.1  &0.1 \\
    batch size &2  &2 \\
    gradient accum. &1 &1 \\
    warmup     &20,000 &20,000 \\
    valid step &20,000 &20,000 \\
    \midrule
    loss$^*$      &\multicolumn{2}{r}{$\mathcal{L}_\text{seq2seq}$ \& $\mathcal{L}_\text{ext}$} \\
    compute   &\multicolumn{2}{r}{{1$\times$GTX TITAN X (12GiB)}} \\
    \bottomrule
  \end{tabular}}
  \caption{RNN Training Hyperparameters. $^*$Both loss functions are cross entropy based.}
\end{table}

\begin{table}[!ht]
  \centering
    \scalebox{0.9}{
  \begin{tabular}{lr}
    \toprule
    Parameter &Value\\
    \midrule
    beam width &4 \\
    length penalty &2.0 \\
    min length &56 \\
    max length$^*$ &144 \& 400 \\
    no repeat trigram size &3 \\
    \bottomrule
  \end{tabular}}
  \caption{Inference Hyperparameters. $^*$144 for Podcast, and 400 for arXiv/PubMed.}
\end{table}

\subsection{Evaluation}
Our ROUGE \cite{lin-2004-rouge} scoring tool is \texttt{pyrouge}, which is a wrapper for perl script.\footnote{\url{https://pypi.org/project/pyrouge/}}

\begin{table*}[!ht]
  \centering
  \scalebox{0.85}{
  \begin{tabular}{lccccc}
    \toprule
    System &Attn-Width &CS-train &Podcast &arXiv &PubMed\\
    \midrule
    BART(1k)   &Full &Truncate              &2.767 &2.179 &1.867 \\
    LoBART(4k) &1024 &Truncate              &2.680 &1.878 &1.530 \\
    $^*$LoBART(4k) &1024 &ORC$_\text{pad-rand}$ &2.647 &1.721 &1.474 \\
    \bottomrule
  \end{tabular}}
  \caption{Performance measured by the average cross-entropy on validation set. $^*$Best system on the test set.}
  \label{tab:valid_performance}
\end{table*}

\begin{table*}[!t]
  \centering
\scalebox{0.85}{
  \begin{tabular}{lcc|ccc|ccc}
    \toprule
    \multirow{2}{*}{System} &\multirow{2}{*}{CS-train} &\multirow{2}{*}{CS-test}  &\multicolumn{3}{c}{arXiv}   &\multicolumn{3}{c}{PubMed} \\
             & &         &R1 &R2 &RL    &R1 &R2 &RL \\
    \midrule
  BART(1k) &\xmark &\xmark &44.96 &17.25 &39.76    &45.06 &18.27 &40.84 \\
  BART(1k) &\xmark &MCS    &46.11 &18.79 &40.83  &46.46 &19.54 &41.91 \\
  BART(1k) &ORC &\xmark    &42.03 &15.62 &37.15 &43.20 &17.02 &39.19 \\
  BART(1k) &ORC &MCS &47.68 &19.77 &42.25    &46.49 &19.45 &42.04 \\
  \midrule
  LoBART(4k) &\xmark &\xmark          &46.90 &18.88 &41.50  &47.40 &20.43 &42.95 \\
  LoBART(4k) &\xmark &MCS     &48.05 &20.11 &42.58  &47.76 &20.76 &43.27 \\
  LoBART(4k) &ORC &\xmark    &46.59 &18.72 &41.24  &47.47 &20.47 &43.02 \\
  LoBART(4k) &ORC &MCS &\textbf{48.79} &\textbf{20.55} &\textbf{43.31}  &\textbf{48.06} &\textbf{20.96} &\textbf{43.56} \\
    \bottomrule
      \end{tabular}
}
  \caption{Extended results on arXiv and PubMed (in Table \ref{tab:arxiv_pubmed_result}). ORC is ORC$_\text{pad-rand}$ training-time content selection.}
  \label{tab:arxiv_pubmed_result_more_results}
\end{table*}

\begin{table*}[!ht]
  \centering
\scalebox{0.85}{
  \begin{tabular}{ll|ccc}
    \toprule
 System  &Initialization   &R1 &R2 &RL \\
\midrule
\multirow{3}{*}{BART(1k)}  &Random  &14.61 &0.82 &11.54 \\
                    &BART-large  &25.82 &9.07 &17.99 \\
                    &BART-large-CNNDM &26.43 &9.22 &18.35 \\
\bottomrule
  \end{tabular}
}
  \caption{Podcast results. The impact of transfer learning. Truncation is applied at both training and test stages.}
  \label{tab:transfer_learning_cnndm_podcast}
\end{table*}

\begin{table*}[!ht]
  \centering
\scalebox{0.85}{
  \begin{tabular}{lll|ccc}
    \toprule
 System &CS-train  &Initialization   &R1 &R2 &RL \\
\midrule
LED(4k)      &Truncate  &$^*$BART-large  &44.40 &17.94 &39.76 \\
\midrule
\multirow{4}{*}{LoBART(4k)}&Truncate  &BART-large                &46.17 &17.96 &40.74 \\
&Truncate  &BART-large-CNNDM          &46.90 &18.88 &41.50 \\
&ORC$_\text{pad-rand}$  &BART-large         &45.25 &17.40 &39.96 \\
&ORC$_\text{pad-rand}$  &BART-large-CNNDM   &46.59 &18.72 &41.24 \\
\bottomrule
  \end{tabular}
}
  \caption{arXiv results. The impact of transfer learning on initializing LoBART. At test time, there is \textit{no} content selection. $^*$To our understanding, LED-large was initialized from BART-large as described in \citet{beltagy2020longformer}.}
  \label{tab:transfer_learning_cnndm}
\end{table*}

\section{Additional Results}
\label{appendix:additional_results}
\noindent \textbf{Losses on Validation Sets}\\
In Table \ref{tab:valid_performance}, we show the standard cross entropy losses on validation sets of our BART/LoBART.

\vspace{4pt}
\noindent \textbf{BART and LoBART on arXiv/PubMed} \\
In Table \ref{tab:arxiv_pubmed_result_more_results}, we provide configurations in addition to Table \ref{tab:arxiv_pubmed_result}. These results (as well as Podcast results in Table \ref{tab:combine_podcast}) show that: in all settings, applying MCS at test time yields a performance gain; and with ORC applied at training, a larger gain is observed.

\vspace{4pt}
\noindent \textbf{Transfer Learning from CNN/DailyMail} \\
In Table \ref{tab:transfer_learning_cnndm_podcast}, we show the impact of transfer learning on fine-tuning BART to Podcast. In Table \ref{tab:transfer_learning_cnndm}, LED(4k) should be very close to LoBART(4k)-TRC-BART-large, we believe that the performance difference is due to the stochastic nature of training. Nevertheless, our experiments are carried out using the same training setting, e.g. hyperparameters, optimizer, etc. Thus, based on the results, we believe that there is an observable improvement due to transfer learning from CNNDM.

\vspace{3pt}
\noindent \textbf{Fine-grained analysis on Podcast test set} \\
\vspace{-13pt}
\begin{figure}[!h]
    \centering
      \includegraphics[width=0.8\linewidth,keepaspectratio]{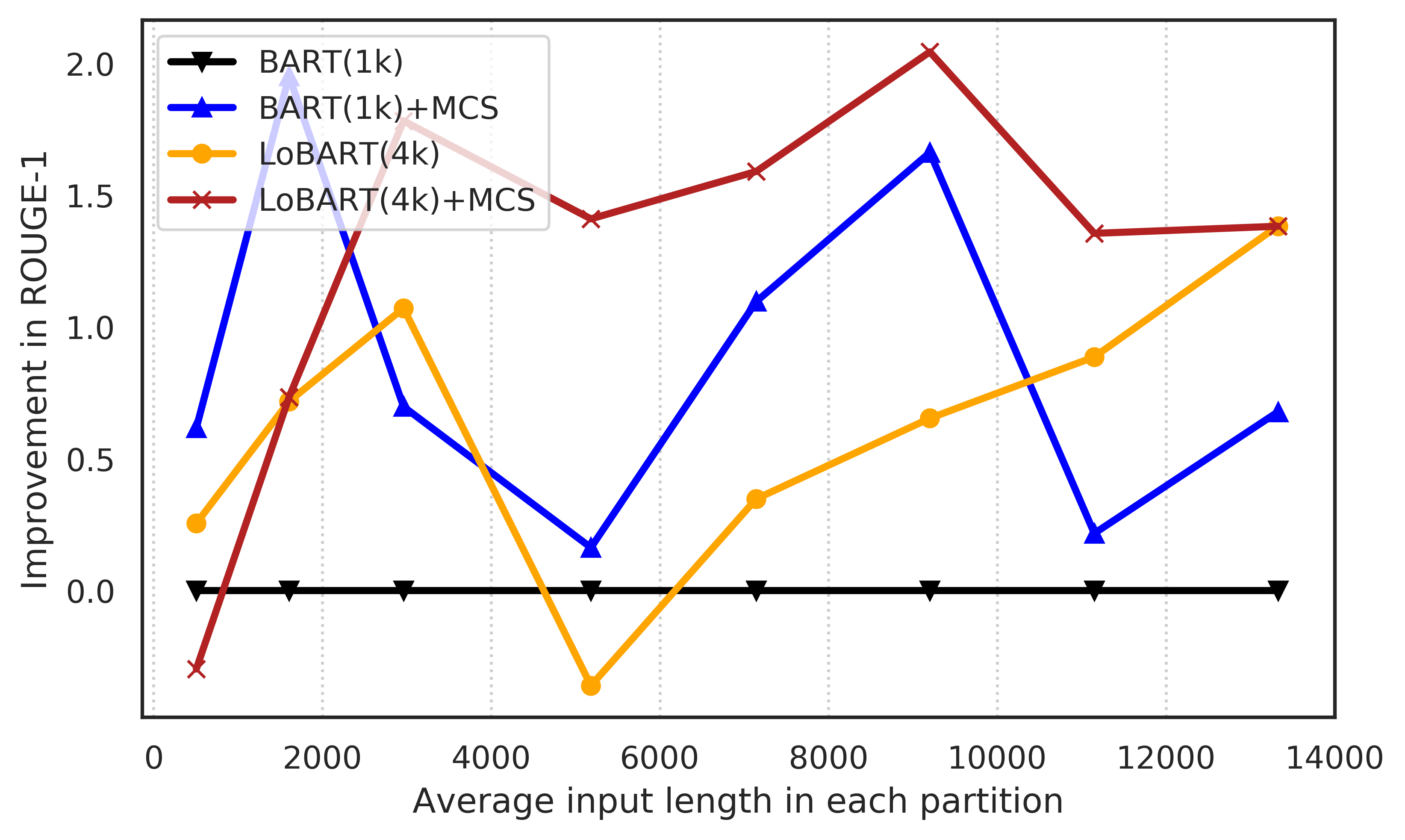}
    \caption{ROUGE-1 score relative to that of BART(1k) on Spotify Podcast (Len:Avg=5,727, 90$^{\text{th}}$\%=11,677).}
    \label{fig:podcast_fine_grained}
\end{figure}


\begin{table*}[ht]
  \centering
  \scalebox{0.9}{
  \begin{tabular}{p{\linewidth}}
    \toprule
    \textbf{Reference Summary}: This week, Irwin and I discuss the iconic designs of the Rolex Submariner and the Porsche 911. Remaining subjectively unchanged through the years. We talk about the subtle changes over the years for this special car and watch and what are the similarities from their history. If you love cars and watches you'll enjoy the show. Grail Watch and Grail Porsche:   Irwin: 1959 Rolex Submariner 5512 x 1967 Porsche 911s   Anthony: 2020 Moser Streamliner Chronograph x 1988 Porsche 911 959.\\
    \midrule
    \textbf{LoBART(4k)+MCS}: In this episode we talk to @grandobsessionsirwin about his passion for Porsche and vintage Rolex. We talk about the Submariner and 911 Submariners, the design evolution over the decades and the future of Porsche in the world of watchmaking. We also chat about the evolution of the Rolex submariner from the 1950s to the present day. \\
    \bottomrule
  \end{tabular}}
  \caption{Spotify Podcast test set sample \#31.}
  \label{tab:spotify_example}
\end{table*}

\begin{table*}[ht]
  \centering
    \scalebox{0.9}{
  \begin{tabular}{p{\linewidth}}
    \toprule
    \textbf{Reference Summary}: we present data from our investigation of the anomalous orange - colored afterglow that was seen in the gammev chameleon afterglow search ( chase ) .
these data includes information about the broad band color of the observed glow , the relationship between the glow and the temperature of the apparatus , and other data taken prior to and during the science operations of chase .
while differing in several details , the generic properties of the afterglow from chase are similar to luminescence seen in some vacuum compounds .
contamination from this , or similar , luminescent signatures will likely impact the design of implementation of future experiments involving single photon detectors and high intensity light sources in a cryogenic environment .\\
    \midrule
    \textbf{LoBART(4k)+MCS}: the gammev chameleon afterglow search ( chase ) experiment at the fermilab tevatron reported the discovery of an anomalous afterglows in its apparatus after shining a high - power pulsed laser into the bore of a cryogenic vacuum chamber immersed in a magnetic field .
we present all of our data that pertains materially to the characterization of the `` orange glow `` signal .
we do not claim any specific explanation of the source or cause of the orange glow , though the dependence upon temperature suggests strongly that the effect is due to some chemical or material property that is excited by the input laser .
the data and discussion presented here may be useful for the design of future experiments that use high intensity light sources in conjunction with single photon detectors in cryogenic environments .\\
    \bottomrule
  \end{tabular}}
  \caption{arXiv test set sample \#315.}
\end{table*}

\begin{table*}[ht]
  \centering
    \scalebox{0.9}{
  \begin{tabular}{p{\linewidth}}
    \toprule
    \textbf{Reference Summary}: the survey of how canadian intensive care units ( icus ) prevent and diagnose venous thromboembolism ( vte ) presented in this issue of critical care illustrates considerable variability .
lack of optimal patient care reflects how vte is rated in icus .
the discussion should no longer focus on the incidence of thrombosis , but rather on its prevention .
unfractionated heparin remains the most commonly used agent to prevent vte , despite the recognized efficacy and safety of low - molecular - weight heparins ( lmwhs ) in the icu setting . in addition , too few icu directors consider the use of mechanical prophylactic measures , such as graded elastic stockings and venous foot pump .
the present situation calls for large randomized controlled trials in either medical or surgical icu patients , and for new education programmes in order to modify the care of icu patients with regard to vte .\\
    \midrule
    \textbf{LoBART(4k)+MCS}: deep vein thrombosis ( dvt ) remains an underestimated problem in intensive care unit ( icu ) patients , despite the findings of many randomized controlled trials performed in the field of dvt prophylaxis after surgery during the past few decades .
the canadian survey reported in the present issue of critical care provides a useful snapshot of daily clinical practice in canada with regard to dvt prevention in icu patients .
it strongly suggests that studies dedicated to this topic should be performed in order to develop useful recommendations .
furthermore , a great effort should be made to educate physicians regarding both dvt screening and pharmacological aspects . \\
    \bottomrule
  \end{tabular}}
  \caption{PubMed test set sample \#3150.}
\end{table*}

\end{document}